\documentclass[dvipsnames]{article}

\usepackage[final]{openbmb_2025}

\usepackage{amsmath}
\usepackage{amssymb}
\usepackage{mathtools}
\usepackage{amsthm}
\tcbuselibrary{breakable}
\usepackage{multirow}
\usepackage{makecell}
\usepackage{subcaption}
\usepackage{wrapfig}
\usepackage{graphicx}
\usepackage{pifont}
\usepackage{tabularx}
\usepackage[utf8]{inputenc} 
\usepackage[T1]{fontenc}    
\usepackage{url}            
\usepackage{booktabs}       
\usepackage{amsfonts}       
\usepackage{nicefrac}       
\usepackage{microtype}      
\usepackage[table]{xcolor}
\definecolor{cellHighlight}{RGB}{235,245,255}
\usepackage{algorithm}
\usepackage{algpseudocode}
\usepackage{color, soul}
\usepackage[most,breakable]{tcolorbox}
\usepackage{xcolor}

\usepackage[labelfont=bf]{caption}
\usepackage{enumitem}
\usepackage{sectsty}
\usepackage{titlesec}
\titlespacing{\paragraph}{0pt}{0.5em}{0.6em}

\usepackage{fancyhdr}
\renewcommand{\headrulewidth}{1pt}
\makeatletter
\def\headrule{{\if@fancyplain\let\headrulewidth\plainheadrulewidth\fi
\hrule\@height\headrulewidth\@width\textwidth \vskip-\headrulewidth}}
\makeatother

\definecolor{BMBDarkBlue}{HTML}{315EFE}
\definecolor{BMBLightBlue}{HTML}{00D3ED}
\usepackage[colorlinks, linkcolor=BMBDarkBlue, anchorcolor=BMBDarkBlue, citecolor=BMBDarkBlue, urlcolor=BMBDarkBlue]{hyperref}

\sectionfont{\color{BMBDarkBlue}\fontfamily{zi4}\selectfont}
\subsectionfont{\color{BMBDarkBlue}\fontfamily{zi4}\selectfont}
\subsubsectionfont{\color{BMBDarkBlue}\fontfamily{zi4}\selectfont}

\newtcolorbox{mytheorem}{
  colback=gray!5,       
  colframe=gray!80,     
  boxrule=0.5pt,        
  arc=4pt,              
  left=4pt,             
  right=4pt,            
  top=4pt,              
  bottom=4pt,           
}

\newcommand{\fancyheadname}{\textit{\textbf{UltraX}}}

\title{UltraX: Refining Pre-Training Data at Scale with\\ Adaptive Programmatic Editing}

\author{
Xinlong Zhao\textsuperscript{1}\thanks{Equal contribution.},
Dongsheng Liu\textsuperscript{2}\footnotemark[1],
Hengyu Zhao\textsuperscript{2},
Zixuan Fu\textsuperscript{3},
Zheng Wang\textsuperscript{3},
Jie Cai\textsuperscript{2},\\
\textbf{
Jie Zhou\textsuperscript{2},
Qiang Ma\textsuperscript{3},
Xuanhe Zhou\textsuperscript{4},
Xu Han\textsuperscript{3},
Yudong Wang\textsuperscript{3}\thanks{Corresponding authors.},
Zhiyuan Liu\textsuperscript{3}\footnotemark[2]
}\\
\textsuperscript{1}Peking University \quad
\textsuperscript{2}ModelBest Inc. \quad
\textsuperscript{3}Tsinghua University \quad
\textsuperscript{4}Shanghai Jiao Tong University \\
\texttt{xlzhao25@stu.pku.edu.cn}
\quad
\texttt{\{wangyudong,liuzy\}@tsinghua.edu.cn}
}

\usepackage{xspace}
\newcommand{\huggingface}{\raisebox{-1.5pt}{\includegraphics[height=1.05em]{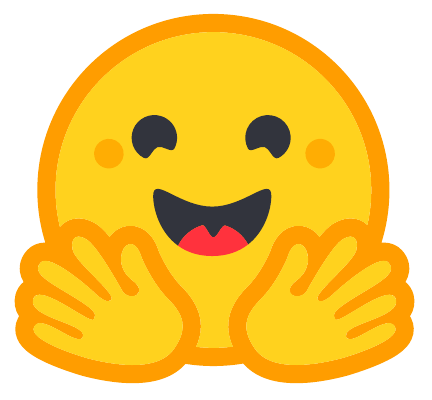}}\xspace}
\newcommand{\github}{\raisebox{-1.5pt}{\includegraphics[height=1.05em]{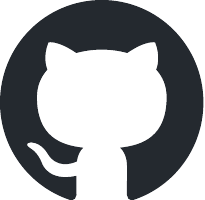}}\xspace}

\begin{document}

\maketitle
\thispagestyle{fancy} 

\vspace{-2.7em}
\begin{center}
\small
\begin{tabular}{l}
    \huggingface\ \textbf{Datasets}: \url{https://huggingface.co/datasets/openbmb/UltraX-Preview} \\
    \huggingface\ \textbf{Model}: \url{https://huggingface.co/openbmb/UltraX-0.6B-Preview} \\
    \github\ \textbf{Code}: \url{https://github.com/openbmb/UltraX}
\end{tabular}
\end{center}
\vspace{-0.4em}

\begin{abstract}
As available training data approaches its physical limit, the performance gains derived from Scaling Laws have begun to diminish. Consequently, the key to further enhancing the performance of Large Language Models (LLMs) has shifted from mere data expansion to improving data utilization efficiency, by enhancing data quality to better exploit the latent potential of existing data. 
However, in the context of large-scale corpora, existing refinement methodologies face significant limitations in quality, efficiency, and reliability: Rule-based approaches are constrained by fixed heuristics and struggle with instance-level variations; LLM-based approaches improve quality but fail to meet the efficiency and reliability requirements of large-scale data processing. 
To address these challenges, we propose UltraX, a function-calling refinement framework for large-scale pre-training data that completes the editing function space by introducing insertion in addition to deletion and modification, enabling fine-grained instance-level editing.
Specifically, UltraX builds a reliable program-supervision generation pipeline. In this pipeline, dataset-adaptive prompt optimization first guides an expert LLM to produce high-quality end-to-end refined texts, and Line Alignment Mapping and Dynamic Context Replacement then convert original-refined text pairs into structured program supervision. 
Meanwhile, UltraX improves supervision quality and stabilizes the training distribution with low-confidence example filtering and ratio-controlled sampling by operation combination. During inference and execution, it normalizes and validates model outputs through sliding-window prediction, global operation aggregation, and systematic post-processing, improving the stability and reliability of large-scale execution. 
Experiments with 1B models pretrained from scratch on multiple corpora show that UltraX achieves the highest average performance across all corpora, with relative improvements exceeding 2\% on multiple datasets. UltraX also matches or surpasses baselines with fewer training tokens, demonstrating stronger data efficiency and refinement reliability.


\end{abstract}

\vspace{-2mm}
\section{Introduction}
\vspace{-2mm}
Scaling laws~\citep{scaling-law} demonstrate that while the performance of Large Language Models (LLMs)~\citep{gpt-4, gemini2.5, hu2024minicpm, minicpm4, qwen3.5, glm5, deepseekv4} continues to improve with the expansion of model parameters and training data, this growth potential has begun to reach a point of diminishing returns as available training data approaches its physical limit. 
Consequently, the focus for further enhancing LLM performance should shift from mere data scaling to the systematic improvement of training data quality~\citep{wang2026data}.

In large-scale corpora, existing data quality improvement methods mainly fall into rule-based filtering and cleaning~\citep{c4, fineweb, redpajama, gopher}, and model-based selection~\citep{ccnet, ultra-fineweb} and refinement~\citep{phi1, wettig2024qurating}. Rule-based methods rely on manually designed heuristics and are computationally efficient, making them widely used in corpus construction. However, their fixed coverage and dependence on expert tuning limit their ability to handle instance-level variation~\citep{zhang2024map, prox}. Model-based methods can leverage learned quality signals for fine-grained data selection or directly revise low-quality texts through refinement, but lightweight models often capture only surface-level statistics and struggle to reliably assess or repair deeper semantic quality issues. 
Recent work further leverages LLMs with stronger understanding and generation capabilities for semantic filtering or end-to-end text refinement~\citep{yu2024mates, dubey2024llama3, li2023textbooks-phi1.5, benallal2024cosmopedia}. However, these methods usually incur substantial computational cost and low throughput, making them difficult to scale to pre-training corpora. 
Overall, existing methods struggle to simultaneously satisfy the demands of semantic-level refinement quality and pre-training-scale processing efficiency. This tension highlights a key challenge: how to retain the benefits of model-driven refinement while reliably scaling data refinement to large pre-training corpora.

We observe that while a performance gap exists between LLMs and lightweight language models in general capabilities, lightweight models can still provide sufficient refinement capability at much lower inference cost when the task is formulated as a structured and task-specific prediction problem. Based on this observation, we explore using lightweight models as efficient refiners for large-scale data processing. 
Crucially, rather than adopting an end-to-end (E2E) text generation method, we implement refinement through predefined function calls. This choice is motivated by two primary considerations: First, E2E generation typically necessitates the production of text at a scale proportional to the input, which entails high inference overhead and reduced throughput. Second, in long-context scenarios, text must inevitably be segmented and reassembled; however, E2E generation struggles to precisely delineate segmentation points and recombination boundaries, which frequently introduces semantic fragmentation and structural inconsistencies~\citep{liu2024best, maini2024rephrasing}. 
In contrast, methods based on function calls output only structured operation sequences, significantly minimizing the output token count. Furthermore, this method is naturally compatible with segment-wise processing and result aggregation, ensuring better coherence when handling long-form documents. 
ProX~\citep{prox} and RefineX~\citep{refinex} are representative methods for function-calling data refinement. Both leverage lightweight models through function calls to achieve data refinement. However, ProX and RefineX exhibit several critical limitations: their function spaces are incomplete and cannot fully model the synergy among insertion, deletion, and modification; their seed supervision either depends on programs directly generated by LLMs or is constrained by deletion-only edit rules, weakening reliability; and their execution procedures still suffer from duplicate string matching, interference between consecutive operations, and insufficient cross-window modeling. 
Therefore, scaling function-calling refinement to large-scale pre-training corpora requires a more complete function space, a more reliable supervision construction pipeline, and a more robust program execution mechanism.

To address these challenges, we introduce \textit{\textbf{UltraX}}, a function-calling refinement framework for large-scale pre-training data. To overcome the incomplete function space, UltraX introduces insertion in addition to deletion and modification, completing the editing function space and enabling fine-grained instance-level editing. To improve seed supervision, UltraX builds a reliable program-supervision generation pipeline: dataset-adaptive prompt optimization first guides an expert LLM to generate high-quality end-to-end refined texts, and Line Alignment Mapping and Dynamic Context Replacement methods then convert original-refined text pairs into structured program supervision. Low-confidence example filtering and ratio-controlled sampling by operation combination are further incorporated to improve supervision quality and stabilize the training distribution. In addition, during inference and execution, UltraX further improves the stability and reliability of large-scale execution by using sliding-window prediction, global operation aggregation, and systematic post-processing to normalize and validate model outputs.

To evaluate the efficacy of UltraX, we conduct from-scratch pre-training experiments with 1B-parameter MiniCPM models~\citep{minicpm4} with 20B-token training budget on five corpora: FineWeb~\citep{fineweb}, RedPajama-V2~\citep{redpajama}, AICC~\citep{ma2025aiccparsehtmlfiner}, Ultra-FineWeb~\citep{ultra-fineweb}, and FineWeb-ProX-Doc~\citep{prox}. 
Experimental results show that UltraX achieves the highest average performance across all corpora, with relative improvements exceeding 2\% on multiple datasets. Moreover, UltraX matches or surpasses baseline methods with fewer training tokens, demonstrating stronger data efficiency and reliability. Further analyses suggest that these gains come from its fine-grained refinement, which better balances noise removal and information preservation.

In summary, the contributions of this paper are as follows:
\begin{itemize}[leftmargin=*]
    \item We propose UltraX, a function-calling refinement framework for large-scale pre-training data, which introduces insertion in addition to deletion and modification to build a more complete editing function space for fine-grained instance-level editing.
    \item We develop a reliable seed-supervision construction pipeline with hierarchical text-to-operation mapping, where dataset-adaptive prompt optimization guides an expert LLM to produce high-quality end-to-end refined texts, and Line Alignment Mapping and Dynamic Context Replacement convert original-refined text pairs into structured program supervision. Low-confidence filtering and ratio-controlled sampling further improve supervision quality and stabilize the training distribution.
    \item We design a robust large-scale execution pipeline with systematic post-processing, including ambiguous replacement filtering, adjacent operation merging, and repetitive pattern fallback, addressing duplicate matching and interference between consecutive operations.
    \item Experimental results show that UltraX consistently improves downstream performance in large-scale pre-training experiments across multiple corpora, while achieving stronger data efficiency and refinement reliability.
\end{itemize}
\vspace{-2mm}
\section{Related Work}
\vspace{-2mm}
\paragraph{Rule-Based Filtering and Cleaning.} Rule-based filtering methodologies predominantly execute quality control at the document level, typically relying on expert-designed heuristics such as URL blacklisting, language identification, gibberish character ratios, and duplication thresholds to prune low-quality documents~\citep{smith2022using, zhang2024map, dou2024sailor, qiu2024wanjuan}. Due to their computational parsimony and ease of implementation, these methods are ubiquitously employed in the construction of massive corpora. However, such rules typically lack the expressive capacity to model instance-level nuances, frequently resulting in the over-zealous rejection of entire documents that may still contain salvageable, high-value information.

In contrast, rule-based cleaning methodologies concentrate on intra-document content restoration. These also depend on manual heuristics to perform local modifications, such as normalizing erratic line breaks, removing hyperlinks, clearing anomalous Base64 encodings, or excising paragraphs lacking proper punctuation~\citep{refinedweb, gopher, fineweb, soldaini-etal-2024-dolma}. While these operations partially mitigate local noise that document-level filtering misses, their efficacy remains constrained by the intrinsic limitations of fixed rule sets and human intuition, rendering them inadequate for capturing increasingly heterogeneous noise patterns. Distinguishing itself from these coarse-grained approaches, UltraX transcends binary document selection. Instead, it facilitates fine-grained refinement within the document through a comprehensive suite of insertion, deletion, and modification operations, thereby unlocking further performance gains.

\paragraph{Model-Based Filtering and Refinement.} With increasing demands for data quality, rule-based methods have become insufficient to meet current requirements. To address this issue, model-based data filtering strategies have gradually become an effective approach to improving data quality in recent years. Traditional quality filtering techniques train classifiers to select high-quality samples~\citep{fineweb, cci3, dclm, chinese_fineweb}. In addition, data filtering methods based on perplexity~\citep{scaling_data, ccnet}, and strategies that use LLMs to evaluate multiple dimensions of data quality through prompts~\citep{how_to, wettig2024qurating}, have been introduced. Nevertheless, these approaches are often hampered by limited robustness, evaluation bias, and high computational cost. Crucially, filtering-centric strategies inherently focus on the binary decision of whether to retain or discard a document, rather than enhancing the intra-document quality of texts that are partially useful yet noisy. 
In contrast, LLM-based refinement methodologies focus more on the direct editing or rewriting of existing data to systematically enhance its quality~\citep{fan2024reformatted, yue2024mammoth2, gunasekar2023textbooks-phi1, li2023textbooks-phi1.5}. However, despite their effectiveness, these methods generally necessitate the generation of text at a volume comparable to the original input, resulting in massive computational overhead that hinders their scalability to pre-training datasets. Furthermore, they inevitably inherit the intrinsic biases and flaws of the underlying generative models and remain susceptible to persistent issues such as hallucinations~\citep{liu2024best}.

To reconcile the trade-off between efficiency and reliability, recent studies have introduced programmatic refinement frameworks leveraging LLMs, such as ProX~\citep{prox} and RefineX~\citep{refinex}. Diverging from direct text rewriting, these methodologies task the model with generating interpretable editing operations that are subsequently applied to the original text, thereby facilitating more granular quality control. However, the effectiveness of such methods heavily depends on the design of editing programs, the quality of seed data construction, and the reliability of post-processing pipelines. Existing approaches still exhibit notable limitations in these aspects, and these challenges further motivate the design of UltraX. Specifically, in terms of operation design, ProX supports only replacement and deletion, while RefineX is restricted to deletion alone; neither provides a complete set of editing operations, limiting the model's ability to restore missing semantics and frequently causing semantic fragmentation. Regarding seed data construction, ProX relies on LLM-generated refinement programs with limited noise coverage, yielding seed data of uncertain quality; RefineX extracts operations from E2E refined text via minimum edit distance, but achieves its deletion-only design by hard-filtering all non-deletion operations. Such coarse hard-filtering fails to capture the synergy among insertion, deletion, and modification. For instance, fixing erroneous segments often requires jointly deleting the original and inserting a more coherent expression, and excluding insertion and modification risks semantic collapse. In terms of execution, ProX lacks a mechanism for disambiguating duplicate substrings within the same document, which can cause edit operations to be applied at unintended positions; moreover, its non-sliding window chunking strategy disrupts semantic continuity across chunk boundaries. RefineX introduces execution-time validation to address some of these issues, yet it overlooks mutual interference among consecutive intra-line operations, which in turn degrades the overall fidelity of the refinement process.
\vspace{-2mm}
\section{Methodology}
\vspace{-2mm}
To address the aforementioned challenges, we introduce UltraX. This section first presents the overall workflow of UltraX, followed by detailed descriptions of the task definition, function space design, program-supervision generation, model training, and inference and execution mechanisms.

\begin{figure}[htbp]
  \centering
  \includegraphics[width=\textwidth]{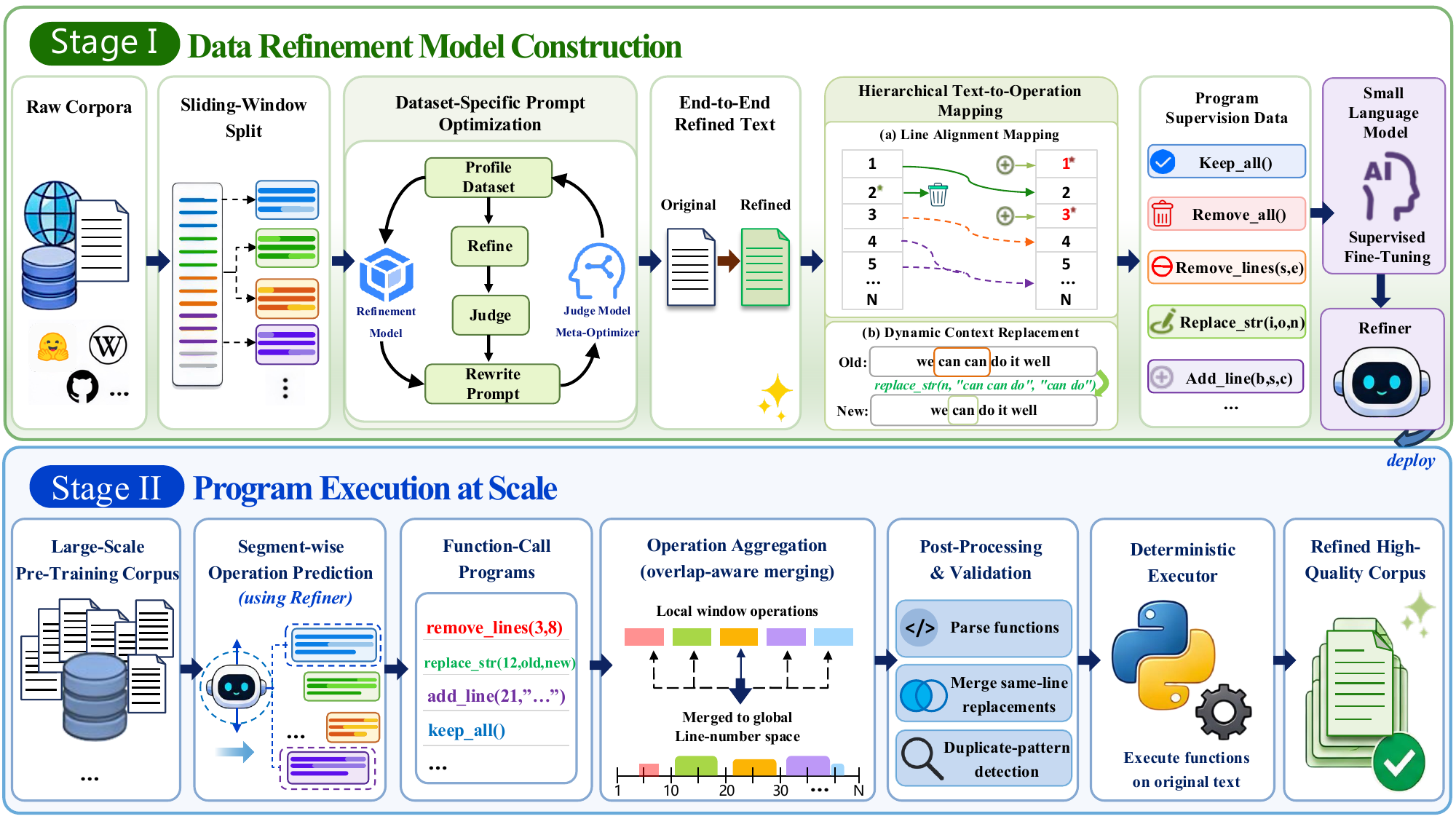}
    \caption{Overall workflow of UltraX, covering program-supervision generation, refinement model training, and inference-time program execution.}
  \label{fig:ultrax_pipeline}
\end{figure}

\vspace{-2mm}
\subsection{Overall Workflow}
\vspace{-2mm}
As shown in Figure~\ref{fig:ultrax_pipeline}, UltraX consists of two stages: refinement model construction and program execution at scale. In Stage I, UltraX starts from raw web corpora, performs seed sampling and sliding-window splitting, and uses dataset-adaptive prompt optimization to guide an expert LLM to generate high-quality end-to-end refined texts. It then applies Line Alignment Mapping followed by Dynamic Context Replacement to convert original-refined text pairs into structured program supervision for supervised fine-tuning of a lightweight refiner. In Stage II, the refiner predicts segment-wise function-call programs over large-scale pre-training corpora; these programs are aggregated across overlapping windows, post-processed and validated, and finally applied by a deterministic executor to produce refined high-quality corpora.

\vspace{-2mm}
\subsection{Task Definition and Program Design}
\vspace{-2mm}
\paragraph{Refinement Task Definition.}
Given an arbitrary document $d \in \mathcal{D}$ from a large-scale pre-training corpus, UltraX formulates data refinement as a programmatic text transformation process. Specifically, we represent the document as a line-organized sequence $d=(l_1,l_2,\ldots,l_n)$, where $l_i$ denotes the $i$-th line of the document. The goal is to learn a refinement model $g_{\theta}$ that generates a structured editing program $\mathcal{Z}=g_{\theta}(d)$ conditioned on the input document, and a deterministic executor $\mathcal{E}$ then applies this program to obtain the refined document $\hat{d}$:
\begin{equation*}
    \hat{d} = \mathcal{E}(\mathcal{Z}, d), \quad \mathcal{Z}=(z_1,z_2,\ldots,z_m), \; z_j \in \Omega .
\end{equation*}
Here, $\Omega$ denotes the function space of UltraX. UltraX formulates refinement as explicit function-call sequences, which avoids high inference overhead and reduces the risk of semantic drift, stylistic rewriting, and structural corruption introduced by generative models~\citep{bi2025parameters}.

\paragraph{Program Function Design.}
To balance execution reliability and inference efficiency, we design a complete program function space $\Omega$, as summarized in Table~\ref{tab:ultrax-program-space}. Unlike function spaces that only support deletion or local modification, UltraX further introduces insertion, allowing the function space to cover the three essential editing behaviors for data refinement: deletion, modification, and insertion. The insertion operation is used to supplement missing structure or restore necessary semantic content. The function space also includes two special document-level operations for keeping or removing the entire input. Specifically, \texttt{keep\_all()} indicates that no modification is required, and \texttt{remove\_all()} indicates that the whole document lacks useful information and should be discarded. The function \texttt{remove\_lines(start\_line, end\_line)} removes a consecutive range of lines, making it suitable for noise such as navigation blocks, advertisements, copyright notices, and template artifacts. The function \texttt{replace\_str(line, source\_str, target\_str)} performs localized string replacement within a specified line, enabling the correction of HTML remnants, inline noise, and lightweight formatting errors. The function \texttt{add\_line(base\_line, sub\_idx, new\_str)} inserts necessary content near a specified position, allowing UltraX to repair missing content introduced by end-to-end refinement or structure restoration. Compared with deletion-only designs, this program space supports a more complete editing process and avoids semantic fragmentation caused by forcibly discarding insertion and replacement operations. At the same time, all operations are anchored by line numbers and local strings rather than raw character indices, which makes program generation easier for lightweight models while preserving compactness, executability, and auditability.

\begin{table}[h]
  \centering
    \caption{Program function definitions in UltraX, designed to compactly cover a complete set of refinement operations.}
    \vspace{1mm}
  \setlength{\tabcolsep}{2.5pt}
  \renewcommand{\arraystretch}{1.1}
  \resizebox{1.0\linewidth}{!}{%
    \begin{tabular}{>{\centering\arraybackslash}p{25em}p{31em}}
    \toprule
    \textbf{Function Interface} & \textbf{Description} \\
    \midrule
    \texttt{keep\_all()} & Return the original text unchanged. \\
    \cmidrule{1-2}
    \texttt{remove\_all()} & Delete the whole text when it lacks useful information. \\
    \cmidrule{1-2}
    \texttt{remove\_lines(start\_line, end\_line)} &
    Delete all content between \texttt{start\_line<int>} and \texttt{end\_line<int>}. \\
    \cmidrule{1-2}
    \texttt{replace\_str(line, source\_str, target\_str)} &
    Replace \texttt{source\_str<str>} with \texttt{target\_str<str>} in \texttt{line<int>}. \\
    \cmidrule{1-2}
    \texttt{add\_line(base\_line, sub\_idx, new\_str)} &
    Insert \texttt{new\_str<str>} near \texttt{base\_line<int>} according to \texttt{sub\_idx<int>}. \\
    \bottomrule
    \end{tabular}%
  }
  \label{tab:ultrax-program-space}
\end{table}

\vspace{-2mm}
\subsection{Data Refinement Model Construction}
\vspace{-2mm}
The goal of this stage is to construct a lightweight data refinement model that directly predicts executable function-call programs from raw text. The overall process consists of the following three steps:

\paragraph{Prompt Optimization and End-to-End Refinement.}
UltraX first samples seed data from multiple web corpora and uses an automatic prompt-optimization agent to analyze the content types and noise patterns of each dataset, producing a refinement prompt adapted to its noise profile. The dataset-adaptive prompt then guides an expert LLM to produce high-quality end-to-end refined text, which serves as the text-level target for program-supervision construction. Details of seed data sampling and prompt optimization are provided in Appendices~\ref{app:seed-sampling-preprocessing} and~\ref{app:prompt-optimization}.

\paragraph{Hierarchical Text-to-Operation Mapping.}
We design a hierarchical mapping process from textual differences to editing operations, progressively converting changes between the original and refined texts into structured function calls. Specifically, Line Alignment Mapping first aligns original and refined lines by considering line content, context, and relative positions, identifying deletions, insertions, and modified lines that require further character-level analysis. Dynamic Context Replacement then analyzes intra-line edit spans at the character level and dynamically expands surrounding context to convert them into uniquely locatable \texttt{replace\_str} operations, thereby avoiding ambiguous matches caused by repeated substrings. In addition, we filter low-confidence examples to improve the reliability of the resulting program supervision. The full construction algorithm and filtering rules are described in Appendix~\ref{app:function-construction}.

\paragraph{Training of the Refinement Model.}
After obtaining structured operation sequences, UltraX converts each raw text into a line-numbered input and uses the corresponding editing sequence as the target output for supervised fine-tuning. To balance the training distribution, we perform ratio-controlled sampling by operation-combination type and analyze its effect in Section~\ref{sec:ablation}. Meanwhile, each training example is augmented with the same system instruction used at inference time, specifying operation definitions, preservation principles, and deletion boundaries. The resulting lightweight refiner only needs to output compact function-call sequences during inference, reducing output token overhead while keeping the results executable and auditable. Training details are provided in Appendix~\ref{app:refiner-training}.

\vspace{-2mm}
\subsection{Program Execution at Scale}
\vspace{-2mm}
After obtaining the refinement model, UltraX formulates large-scale data refinement as program prediction followed by deterministic program execution. This inference-and-execution stage consists of the following three steps:

\paragraph{Segment-wise Operation Prediction.}
For each document to be refined, UltraX first normalizes newline characters and prefixes each line with an explicit line-number marker. For documents that exceed the length limit, we adopt a sliding-window strategy to split the text into segments, as detailed in Appendix~\ref{app:sliding-window-reassembly}. The trained refinement model then independently generates an operation sequence for each segment.

\paragraph{Operation Aggregation and Post-processing.}
Segment-wise prediction produces multiple local operation sequences, which UltraX merges into a global program. Specifically, local operations are mapped back to the global line-number space according to the line range of each sliding-window segment, and only operations from non-duplicated regions are retained to avoid modifying the same text span multiple times. UltraX then parses and normalizes model outputs, filters malformed or unparsable function calls, and handles special cases such as \texttt{keep\_all} and \texttt{remove\_all} in a unified manner. It further filters ambiguous replacements, merges consecutive or mutually interfering \texttt{replace\_str} operations on the same line, and detects abnormal repetitive function patterns, thereby converting candidate programs into stable, compact, and executable global refinement programs. Details of reassembly, post-processing, and duplicate-pattern fallback are provided in Appendices~\ref{app:sliding-window-reassembly}, \ref{app:post-processing}, and~\ref{app:fallback-duplicate-detection}.

\paragraph{Program Execution and Refined Corpus Generation.}
After obtaining the post-processed global program, UltraX applies the editing operations to the original text through a deterministic function executor. Since the predicted operations and executed outputs for each example can be stored and inspected, the refinement process is highly traceable. In this way, UltraX transforms raw corpora into programmatically refined high-quality text at scale, providing an efficient, stable, and reliable refinement mechanism for large-scale pre-training data.

\vspace{-2mm}
\section{Experiments}
\vspace{-2mm}
\subsection{Experimental Setting}
\vspace{-2mm}
\paragraph{Training Corpora and Base Model Selection.}
We use five representative large-scale pre-training corpora as our data sources: FineWeb~\citep{fineweb}, RedPajama-v2~\citep{redpajama}, AICC~\citep{ma2025aiccparsehtmlfiner}, Ultra-FineWeb~\citep{ultra-fineweb}, and FineWeb-ProX-Doc~\citep{prox}. For each corpus, we construct approximately 20B-token training sets for Raw, ProX-C, and UltraX, and pretrain an approximately 1B-parameter MiniCPM model~\citep{minicpm4} from scratch to ensure fair comparison. Data source and sampling details are provided in Appendix~\ref{app:data-sources}, and model architecture and training details are provided in Appendix~\ref{app:model-architecture-training}.

\paragraph{Baselines and Evaluation Setup.}
Since RefineX~\citep{refinex} has not been open-sourced to date, we choose ProX-C~\citep{prox} as the primary baseline. To ensure fair comparison, we follow the open-source settings of ProX-C as closely as possible, including its text segmentation strategy, program function design, and inference configuration. Apart from the data refinement pipeline, all pre-training experiments share the same base model, training scale, and training configuration, so that downstream performance differences can be primarily attributed to the refinement method itself. After pre-training, we evaluate all models with LightEval~\citep{lighteval} on ten benchmarks: the nine ``early signal'' tasks used by FineWeb~\citep{fineweb} plus SciQ~\citep{welbl2017crowdsourcing-sciq}. Benchmark setup and aggregation details are described in Appendices~\ref{app:benchmark-setup} and~\ref{app:sampling-aggregation}, while refined text evaluation details are provided in Appendix~\ref{app:data-quality-eval}.

\vspace{-2mm}
\subsection{Evaluation of Performance on Pre-trained Language Models}
\vspace{-2mm}
\label{sec:main_exp}
In this section, we evaluate the effectiveness of UltraX by pretraining language models from scratch on data produced by different refinement methods. We first report the main results to verify the overall effectiveness of UltraX across heterogeneous corpora. We then analyze the performance dynamics on FineWeb under different training token budgets to examine the data efficiency of UltraX.

\begin{table}[h]
  \centering
    \caption{Performance on ten downstream tasks across five pre-training corpora. \textbf{Bolded} entries denote the best result within each corpus, and \textbf{\#Win} counts the number of tasks where each method achieves the best performance.}
    \vspace{1mm}
  \setlength{\tabcolsep}{2.4pt}
  \resizebox{1.0\textwidth}{!}{%
  \begin{tabular}{ll|cccccccccc|cc}
  \toprule
  \textbf{Corpus} & \textbf{Method} & \textbf{ARC-C} & \textbf{ARC-E} & \textbf{CSQA} & \textbf{HellaS} & \textbf{MMLU} & \textbf{OBQA} & \textbf{PIQA} & \textbf{SIQA} & \textbf{WinoG} & \textbf{SciQ} & \textbf{Avg} & \textbf{\#Win} \\
  \midrule
  \multirow{3}{*}{FineWeb} & Raw & 25.85 & 45.66 & 35.63 & 44.89 & 28.51 & 31.60 & 70.35 & 43.24 & 51.54 & 73.50 & 45.08 & 0 / 10 \\
   & ProX-C & 25.09 & 45.20 & 36.94 & 45.32 & 28.49 & 31.40 & 71.16 & 42.94 & 51.14 & 72.80 & 45.05 & 0 / 10 \\
   & \cellcolor{cellHighlight}UltraX & \cellcolor{cellHighlight}\textbf{26.62} & \cellcolor{cellHighlight}\textbf{45.96} & \cellcolor{cellHighlight}\textbf{37.43} & \cellcolor{cellHighlight}\textbf{46.29} & \cellcolor{cellHighlight}\textbf{28.90} & \cellcolor{cellHighlight}\textbf{31.80} & \cellcolor{cellHighlight}\textbf{71.98} & \cellcolor{cellHighlight}\textbf{43.71} & \cellcolor{cellHighlight}\textbf{52.72} & \cellcolor{cellHighlight}\textbf{76.00} & \cellcolor{cellHighlight}\textbf{46.14} & \cellcolor{cellHighlight}10 / 10 \\
  \midrule
  \multirow{3}{*}{RedPajama-v2} & Raw & 23.46 & 43.69 & 32.02 & 39.64 & 27.50 & 30.80 & \textbf{68.66} & 41.56 & \textbf{52.64} & 70.90 & 43.09 & 2 / 10 \\
   & ProX-C & \textbf{25.60} & 44.23 & 33.01 & 40.19 & 27.63 & 31.00 & 68.61 & 42.37 & 50.43 & 71.60 & 43.47 & 1 / 10 \\
   & \cellcolor{cellHighlight}UltraX & \cellcolor{cellHighlight}24.40 & \cellcolor{cellHighlight}\textbf{45.62} & \cellcolor{cellHighlight}\textbf{33.91} & \cellcolor{cellHighlight}\textbf{40.80} & \cellcolor{cellHighlight}\textbf{27.88} & \cellcolor{cellHighlight}\textbf{32.20} & \cellcolor{cellHighlight}68.44 & \cellcolor{cellHighlight}\textbf{42.68} & \cellcolor{cellHighlight}51.30 & \cellcolor{cellHighlight}\textbf{72.60} & \cellcolor{cellHighlight}\textbf{43.98} & \cellcolor{cellHighlight}7 / 10 \\
  \midrule
  \multirow{3}{*}{AICC} & Raw & 24.06 & 40.99 & 31.70 & 35.91 & \textbf{27.37} & 28.80 & 66.16 & \textbf{42.07} & 49.57 & 69.70 & 41.63 & 2 / 10 \\
   & ProX-C & \textbf{25.17} & 42.09 & \textbf{32.27} & 37.85 & 26.95 & \textbf{30.20} & 66.70 & 41.40 & 49.64 & 69.20 & 42.15 & 3 / 10 \\
   & \cellcolor{cellHighlight}UltraX & \cellcolor{cellHighlight}24.49 & \cellcolor{cellHighlight}\textbf{42.85} & \cellcolor{cellHighlight}31.61 & \cellcolor{cellHighlight}\textbf{38.03} & \cellcolor{cellHighlight}26.98 & \cellcolor{cellHighlight}29.80 & \cellcolor{cellHighlight}\textbf{68.34} & \cellcolor{cellHighlight}41.71 & \cellcolor{cellHighlight}\textbf{50.28} & \cellcolor{cellHighlight}\textbf{70.20} & \cellcolor{cellHighlight}\textbf{42.43} & \cellcolor{cellHighlight}5 / 10 \\
  \midrule
  \multirow{3}{*}{Ultra-FineWeb} & Raw & \textbf{32.00} & 57.28 & 33.58 & 44.56 & 30.62 & 34.60 & 70.40 & 41.15 & \textbf{51.85} & 78.50 & 47.45 & 2 / 10 \\
   & ProX-C & 31.31 & 56.61 & 32.84 & 45.29 & \textbf{31.03} & 35.20 & \textbf{71.65} & \textbf{42.27} & 49.88 & 76.70 & 47.28 & 3 / 10 \\
   & \cellcolor{cellHighlight}UltraX & \cellcolor{cellHighlight}31.31 & \cellcolor{cellHighlight}\textbf{58.25} & \cellcolor{cellHighlight}\textbf{33.91} & \cellcolor{cellHighlight}\textbf{45.66} & \cellcolor{cellHighlight}30.97 & \cellcolor{cellHighlight}\textbf{37.20} & \cellcolor{cellHighlight}70.08 & \cellcolor{cellHighlight}42.17 & \cellcolor{cellHighlight}51.38 & \cellcolor{cellHighlight}\textbf{80.50} & \cellcolor{cellHighlight}\textbf{48.14} & \cellcolor{cellHighlight}5 / 10 \\
  \midrule
  \multirow{3}{*}{FineWeb-ProX-Doc} & Raw & 29.35 & 52.57 & 35.22 & 46.31 & 30.21 & 35.20 & 69.37 & 42.27 & \textbf{52.41} & 76.00 & 46.89 & 1 / 10 \\
   & ProX-C & 29.35 & \textbf{55.05} & 35.22 & 46.41 & 30.40 & 35.60 & \textbf{69.64} & 42.48 & 51.14 & 76.60 & 47.19 & 2 / 10 \\
   & \cellcolor{cellHighlight}UltraX & \cellcolor{cellHighlight}\textbf{30.38} & \cellcolor{cellHighlight}55.01 & \cellcolor{cellHighlight}\textbf{36.53} & \cellcolor{cellHighlight}\textbf{47.17} & \cellcolor{cellHighlight}\textbf{31.05} & \cellcolor{cellHighlight}\textbf{37.00} & \cellcolor{cellHighlight}69.15 & \cellcolor{cellHighlight}\textbf{42.99} & \cellcolor{cellHighlight}51.70 & \cellcolor{cellHighlight}\textbf{78.30} & \cellcolor{cellHighlight}\textbf{47.93} & \cellcolor{cellHighlight}7 / 10 \\
  \bottomrule
  \end{tabular}%
  }
  \label{tab:main-exp-performance}
\end{table}

\paragraph{UltraX Demonstrates Strong Performance across Diverse Pre-Training Corpora.}

\begin{wrapfigure}{r}{0.47\textwidth}
    \centering
    \vspace{-5mm}
    \includegraphics[width=0.45\textwidth]{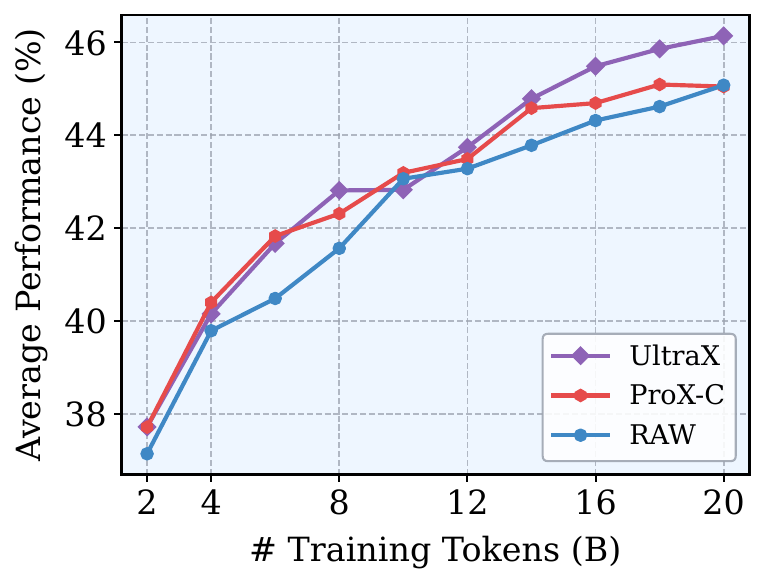}
    \vspace{-2mm}
    \caption{Average downstream performance on FineWeb under different training token budgets.}
    \label{fig:fineweb-token-curve}
\end{wrapfigure}

Table~\ref{tab:main-exp-performance} presents the main experimental results across five pre-training corpora. Overall, UltraX achieves the highest average performance on all five corpora, demonstrating that its refinement capability generalizes consistently across data sources with different quality distributions. Specifically, compared with Raw and ProX-C, UltraX achieves average relative improvements of approximately 2.00\% and 1.53\%, respectively. Beyond average performance, UltraX obtains the best result on 34 out of 50 task-corpus pairs, indicating that its gains are not caused by isolated benchmark fluctuations but are consistently reflected across reasoning, commonsense, and knowledge-oriented tasks. Notably, UltraX still improves over strong corpora such as Ultra-FineWeb and FineWeb-ProX-Doc, which already exhibit high initial quality. This suggests that fine-grained programmatic refinement can further enhance the effective training value of curated corpora. The complete experimental results are provided in Appendix~\ref{app:full-results}.

\newpage
\paragraph{UltraX Demonstrates Better Performance with Fewer Training Tokens.}
Figure~\ref{fig:fineweb-token-curve} shows the average downstream performance of Raw, ProX-C, and UltraX on FineWeb under different training token budgets. The results show that UltraX not only achieves the best final performance but also reaches strong performance earlier during pretraining. Specifically, UltraX attains an average score of 45.49 with only 16B training tokens, already surpassing the final performance of Raw and ProX-C trained with 20B tokens (45.08 and 45.05, respectively). When trained with 20B tokens, UltraX further improves to 46.14. In contrast, as the number of training tokens increases, the performance gains of Raw and ProX-C become relatively saturated in the later stages. These results indicate that UltraX improves the effective information density of training tokens through fine-grained, instance-level programmatic refinement, enabling the model to acquire stronger downstream capabilities with fewer training tokens. Full per-checkpoint and per-benchmark results are reported in Appendix~\ref{app:full-results}.

\vspace{-2mm}
\subsection{Ablation Studies}
\vspace{-2mm}
\label{sec:ablation}
In this section, to further analyze the impact of key design choices in UltraX, we study four aspects: the role of system instructions during refinement model training and inference, the operation distribution of seed data, quality-stratified refinement strategies for high and low-quality subsets, and the contribution of different program function spaces. All ablation experiments are conducted on the 20B token training corpus sampled from FineWeb, using the same 1B MiniCPM architecture and training configuration as in the main experiments. For clarity, we state in advance that, in the main experiments, UltraX adopts the instruction-guided setting by default and uses the edit-weighted seed data sampling strategy. Therefore, the first two ablation studies follow a controlled-variable design: only one key factor is changed at a time, while all other settings are kept identical.

\paragraph{Effect of Instruction-Guided Refinement.}
We first examine the role of system instructions in the training and inference of the refinement model. Table~\ref{tab:ablation-instruction} compares Raw, UltraX trained and inferred without system instructions, and UltraX guided by instruction. The results show that programmatic refinement already brings clear gains even without system instructions, improving the average score from 45.08 to 45.73. However, after introducing system instructions, the model receives clearer operation definitions, preservation principles, and deletion boundaries, further improving the average score to 46.14 and achieving the best result on 8 out of 10 tasks. This demonstrates that an explicit task protocol improves the stability and generalization of refinement program generation. The system instruction is provided in Appendix~\ref{app:function-construction}.

\begin{table}[h]
  \centering
  \caption{Ablation on instruction-guided refinement. Both UltraX variants use the edit-weighted sampling strategy for training data construction.}
  \vspace{1mm}
  \setlength{\tabcolsep}{2.8pt}
  \resizebox{1.0\textwidth}{!}{%
  \begin{tabular}{l|cccccccccc|cc}
  \toprule
  \textbf{Method} & \textbf{ARC-C} & \textbf{ARC-E} & \textbf{CSQA} & \textbf{HellaS} & \textbf{MMLU} & \textbf{OBQA} & \textbf{PIQA} & \textbf{SIQA} & \textbf{WinoG} & \textbf{SciQ} & \textbf{Avg} & \textbf{\#Win} \\
  \midrule
  Raw & 25.85 & 45.66 & 35.63 & 44.89 & 28.51 & 31.60 & 70.35 & 43.24 & 51.54 & 73.50 & 45.08 & 0 / 10 \\
   UltraX (No-Instruction) & 26.02 & \textbf{47.18} & 35.79 & \textbf{46.50} & 28.25 & 31.20 & 71.65 & 43.55 & 52.57 & 74.60 & 45.73 & 2 / 10 \\
  UltraX (Instruction-Guided) & \textbf{26.62} & 45.96 & \textbf{37.43} & 46.29 & \textbf{28.90} & \textbf{31.80} & \textbf{71.98} & \textbf{43.71} & \textbf{52.72} & \textbf{76.00} & \textbf{46.14} & 8 / 10 \\
  \bottomrule
  \end{tabular}%
  }

  \label{tab:ablation-instruction}
\end{table}

\paragraph{Effect of Seed Operation Distribution.}
We next analyze how the operation distribution in seed data affects the refinement model. We compare two sampling strategies: preservation-weighted sampling and edit-weighted sampling, whose detailed operation distributions are shown in Table~\ref{tab:ablation-operation-distribution}. As shown in Table~\ref{tab:ablation-sampling}, both strategies significantly outperform Raw, indicating that learning explicit refinement operations is consistently beneficial. Preservation-weighted sampling performs better on tasks such as ARC-C, ARC-E, OBQA, and SciQ, while edit-weighted sampling achieves the highest average score and performs best on CSQA, MMLU, PIQA, SIQA, and WinoGrande. This suggests that increasing edit-oriented supervision strengthens fine-grained cleaning ability, while retaining an appropriate amount of \texttt{keep\_all} supervision helps prevent over-editing.

\begin{table}[h]
  \centering
    \caption{Distribution of operation-combination categories under the two seed data sampling strategies. Each training example is assigned to exactly one category according to the set of operation types appearing in its target program, and is therefore not counted repeatedly across different categories. Percentages indicate the proportion of training examples in the final sampled SFT data.}
    \vspace{1mm}
  \small
  \setlength{\tabcolsep}{4pt}
  \resizebox{0.78\textwidth}{!}{%
  \begin{tabular}{lcc}
  \toprule
  \textbf{Operation Combination} & \textbf{Preservation-Weighted} & \textbf{Edit-Weighted} \\
  \midrule
  \texttt{keep\_all} & 60.000\% & 14.419\% \\
  \texttt{remove\_lines + replace\_str} & 24.293\% & 45.921\% \\
  \texttt{remove\_lines} & 7.678\% & 15.307\% \\
  \texttt{replace\_str} & 4.870\% & 12.874\% \\
  \texttt{remove\_all} & 2.742\% & 7.535\% \\
  \texttt{add\_line + remove\_lines + replace\_str} & 0.228\% & 2.154\% \\
  \texttt{add\_line + remove\_lines} & 0.152\% & 1.440\% \\
  \texttt{add\_line + replace\_str} & 0.036\% & 0.337\% \\
  \texttt{add\_line} & 0.001\% & 0.013\% \\
  \bottomrule
  \end{tabular}%
  }

  \label{tab:ablation-operation-distribution}
\end{table}

\begin{table}[h]
  \centering
    \caption{Ablation on seed operation distribution. Both UltraX variants are trained and inferred with system instructions.}
    \vspace{1mm}
  \setlength{\tabcolsep}{2.8pt}
  \resizebox{1.0\textwidth}{!}{%
  \begin{tabular}{l|cccccccccc|cc}
  \toprule
  \textbf{Method} & \textbf{ARC-C} & \textbf{ARC-E} & \textbf{CSQA} & \textbf{HellaS} & \textbf{MMLU} & \textbf{OBQA} & \textbf{PIQA} & \textbf{SIQA} & \textbf{WinoG} & \textbf{SciQ} & \textbf{Avg} & \textbf{\#Win} \\
  \midrule
  Raw & 25.85 & 45.66 & 35.63 & 44.89 & 28.51 & 31.60 & 70.35 & 43.24 & 51.54 & 73.50 & 45.08 & 0 / 10 \\
  UltraX (Preservation-Weighted) & \textbf{26.79} & \textbf{47.14} & 36.12 & \textbf{46.30} & 28.54 & \textbf{33.80} & 70.67 & 42.78 & 51.62 & \textbf{76.10} & 45.99 & 5 / 10 \\
  UltraX (Edit-Weighted) & 26.62 & 45.96 & \textbf{37.43} & 46.29 & \textbf{28.90} & 31.80 & \textbf{71.98} & \textbf{43.71} & \textbf{52.72} & 76.00 & \textbf{46.14} & 5 / 10 \\
  \bottomrule
  \end{tabular}%
  }

  \label{tab:ablation-sampling}
\end{table}

\paragraph{Quality-Stratified Refinement.}
To study refinement strategies on subsets with different quality levels, we use the Ultra-FineWeb classifier to partition the 20B token FineWeb corpus, setting the threshold to 0.05. Under this threshold, the high-quality Head subset accounts for approximately 21.5\%, while the low-quality Tail subset accounts for approximately 78.5\%. We conduct two groups of experiments: the first fixes the Head subset as Raw and varies the refinement strategy for Tail, while the second fixes Tail as Raw and varies the refinement strategy for Head.

Table~\ref{tab:ablation-tail} reports the results when Head is fixed as Raw and different strategies are applied to Tail. Refining the Tail subset generally leads to clear improvements, with UltraX (No-Instruction) achieving the highest average score of 45.82 in this group. This indicates that the low-quality subset contains more noise that can be removed or repaired through refinement, making fine-grained Tail refinement more likely to translate into downstream gains. Notably, UltraX (No-Instruction) lacks the preservation constraints imposed by system instructions and therefore tends to perform more aggressive modifications. On the noise-dense Tail subset, such stronger editing tendency can instead lead to better refinement effects. In contrast, the preservation-weighted strategy is weaker on Tail, suggesting that low-quality regions benefit more from active edit-oriented refinement than from conservative preservation.

\begin{table}[h]
  \centering
    \caption{Ablation on Tail refinement strategies with the Head subset fixed as Raw.}
    \vspace{1mm}
  \setlength{\tabcolsep}{2.8pt}
  \resizebox{1.0\textwidth}{!}{%
  \begin{tabular}{l|cccccccccc|cc}
  \toprule
  \textbf{Tail Strategy} & \textbf{ARC-C} & \textbf{ARC-E} & \textbf{CSQA} & \textbf{HellaS} & \textbf{MMLU} & \textbf{OBQA} & \textbf{PIQA} & \textbf{SIQA} & \textbf{WinoG} & \textbf{SciQ} & \textbf{Avg} & \textbf{\#Win} \\
  \midrule
  Raw & 25.85 & 45.66 & 35.63 & 44.89 & 28.51 & 31.60 & 70.35 & \textbf{43.24} & 51.54 & 73.50 & 45.08 & 1 / 10 \\
  ProX-C & 25.43 & 46.68 & 36.36 & 46.05 & 28.37 & 32.80 & 71.27 & 42.43 & 52.49 & 74.60 & 45.65 & 0 / 10 \\
  UltraX (No-Instruction) & \textbf{26.88} & \textbf{47.10} & 35.46 & \textbf{46.17} & 28.34 & \textbf{33.60} & 70.57 & 43.14 & 52.41 & 74.50 & \textbf{45.82} & 4 / 10 \\
  UltraX (Preservation-Weighted) & 26.71 & 44.91 & 35.46 & 45.86 & \textbf{28.88} & 32.20 & 70.89 & 42.12 & \textbf{52.57} & 72.70 & 45.23 & 2 / 10 \\
  UltraX & 25.51 & 46.30 & \textbf{37.59} & 45.77 & 28.18 & 31.80 & \textbf{72.03} & 42.17 & 50.67 & \textbf{75.90} & 45.59 & 3 / 10 \\
  \bottomrule
  \end{tabular}%
  }

  \label{tab:ablation-tail}
\end{table}

Table~\ref{tab:ablation-head} reports the results when Tail is fixed as Raw and different strategies are applied to Head. Compared with Tail refinement, refining only the Head subset yields more limited and less stable gains. UltraX (Preservation-weighted) achieves the highest average score of 45.38, but the improvement is smaller than that obtained by Tail refinement. This is expected because the Head subset is already of higher quality and leaves less room for refinement; aggressive editing may even damage valuable content. Therefore, conservative refinement is more suitable for high-quality data, while the major gains primarily come from refining the low-quality Tail subset.

\begin{table}[h]
  \centering
    \caption{Ablation on Head refinement strategies with the Tail subset fixed as Raw.}
    \vspace{1mm}
  \setlength{\tabcolsep}{2.8pt}
  \resizebox{1.0\textwidth}{!}{%
  \begin{tabular}{l|cccccccccc|cc}
  \toprule
  \textbf{Head Strategy} & \textbf{ARC-C} & \textbf{ARC-E} & \textbf{CSQA} & \textbf{HellaS} & \textbf{MMLU} & \textbf{OBQA} & \textbf{PIQA} & \textbf{SIQA} & \textbf{WinoG} & \textbf{SciQ} & \textbf{Avg} & \textbf{\#Win} \\
  \midrule
  Raw & \textbf{25.85} & 45.66 & 35.63 & 44.89 & \textbf{28.51} & 31.60 & 70.35 & 43.24 & 51.54 & 73.50 & 45.08 & 2 / 10 \\
  ProX-C & 25.77 & \textbf{46.30} & 35.30 & 45.24 & 28.34 & 31.80 & \textbf{70.84} & 42.27 & 50.51 & 73.10 & 44.95 & 2 / 10 \\
  UltraX (No-Instruction) & 25.51 & 44.15 & 35.71 & \textbf{45.46} & 28.27 & 33.20 & 70.02 & 43.14 & 50.99 & 71.20 & 44.77 & 1 / 10 \\
  UltraX (Preservation-Weighted) & 24.49 & 45.75 & \textbf{35.87} & 45.25 & 28.17 & 33.40 & 70.67 & \textbf{43.96} & 51.70 & \textbf{74.50} & \textbf{45.38} & 3 / 10 \\
  UltraX & 25.51 & 46.00 & 35.71 & 45.24 & 28.02 & \textbf{34.20} & 70.18 & 43.45 & \textbf{52.17} & 73.10 & 45.36 & 2 / 10 \\
  \bottomrule
  \end{tabular}%
  }

  \label{tab:ablation-head}
\end{table}

\paragraph{Ablation on Program Function Space.}
Finally, we analyze the effect of different program function spaces in UltraX. Table~\ref{tab:ablation-functions} compares four function-space settings: document-level decisions only (\texttt{keep\_all} and \texttt{remove\_all}), document-level decisions augmented with line-level editing (\texttt{add\_line} and \texttt{remove\_lines}), document-level decisions augmented with intra-line replacement (\texttt{replace\_str}), and the full UltraX function space with all operations. Except for the full UltraX setting, the other variants are obtained by hard-filtering the complete function space to retain specific subsets of functions. Document-level decision alone achieves an average score of 45.22, indicating that keeping or removing whole documents can already filter part of the low-value data. However, adding only intra-line replacement on top of document-level decisions yields the weakest performance, suggesting that without line-level deletion and insertion, the model struggles to handle long-span structural redundancy in web data. Line-level editing brings additional gains, but still underperforms the full UltraX setting. The complete function space achieves the highest average score of 46.14 and wins on 8 out of 10 tasks, further confirming that the synergy among \texttt{remove\_all}, \texttt{remove\_lines}, \texttt{replace\_str}, and \texttt{add\_line} is crucial for reliable data refinement.

\begin{table}[h]
  \centering
    \caption{Ablation on the program function space of UltraX.}
    \vspace{1mm}
  \setlength{\tabcolsep}{2.8pt}
  \resizebox{1.0\textwidth}{!}{%
  \begin{tabular}{l|cccccccccc|cc}
  \toprule
  \textbf{Function Space} & \textbf{ARC-C} & \textbf{ARC-E} & \textbf{CSQA} & \textbf{HellaS} & \textbf{MMLU} & \textbf{OBQA} & \textbf{PIQA} & \textbf{SIQA} & \textbf{WinoG} & \textbf{SciQ} & \textbf{Avg} & \textbf{\#Win} \\
  \midrule
  Document-Level Decision & \textbf{27.30} & 45.58 & 34.40 & 45.59 & 28.04 & 32.20 & 70.46 & \textbf{43.71} & 51.78 & 73.10 & 45.22 & 2 / 10 \\
  Line-Level Editing & 26.37 & 45.24 & 36.20 & 46.18 & 28.38 & \textbf{33.40} & 70.40 & 42.27 & 51.54 & 73.50 & 45.35 & 1 / 10 \\
  Intra-Line Replacement & 24.32 & 45.79 & 36.61 & 45.21 & 28.40 & 31.20 & 70.73 & 41.61 & 50.83 & 74.40 & 44.91 & 0 / 10 \\
  Full UltraX & 26.62 & \textbf{45.96} & \textbf{37.43} & \textbf{46.29} & \textbf{28.90} & 31.80 & \textbf{71.98} & \textbf{43.71} & \textbf{52.72} & \textbf{76.00} & \textbf{46.14} & 8 / 10 \\
  \bottomrule
  \end{tabular}%
  }
  \label{tab:ablation-functions}
\end{table}

\vspace{-2mm}
\subsection{In-Depth Analysis}
\label{sec:in_depth_analysis}
\vspace{-2mm}

In this section, we further analyze UltraX from three perspectives: token count distribution, generated refinement functions, and refined data quality. These analyses provide a more fine-grained explanation of the main experimental results and illustrate how UltraX balances noise removal with information preservation.

\begin{figure}[h]
    \centering
    \begin{minipage}{0.32\textwidth}
        \centering
        \includegraphics[width=\textwidth]{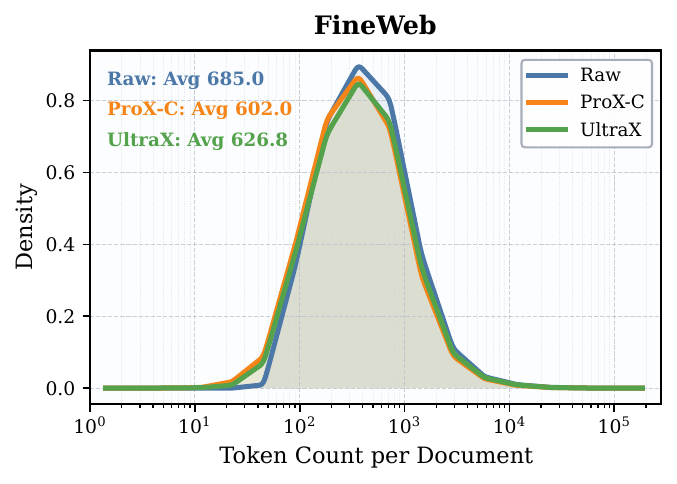}
        \vspace{-3mm}
    \end{minipage}
    \hfill
    \begin{minipage}{0.32\textwidth}
        \centering
        \includegraphics[width=\textwidth]{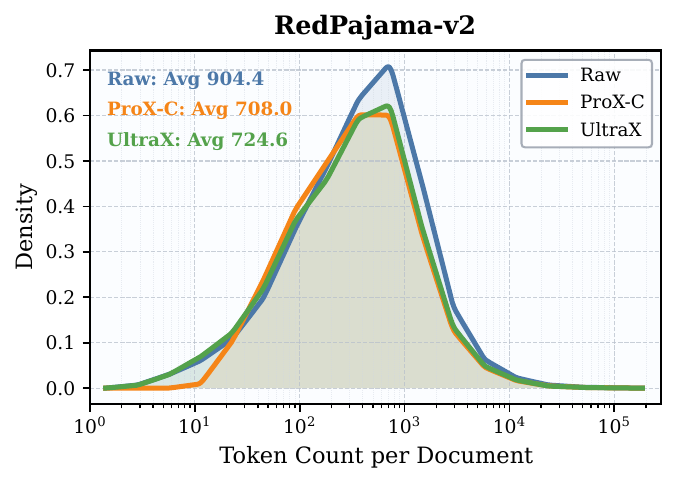}
        \vspace{-3mm}
    \end{minipage}
    \hfill
    \begin{minipage}{0.32\textwidth}
        \centering
        \includegraphics[width=\textwidth]{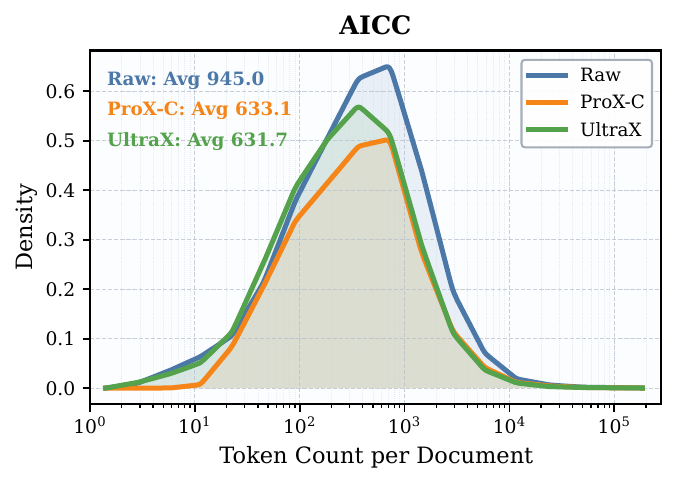}
        \vspace{-3mm}
    \end{minipage}

    \vspace{1mm}

    \begin{minipage}{0.32\textwidth}
        \centering
        \includegraphics[width=\textwidth]{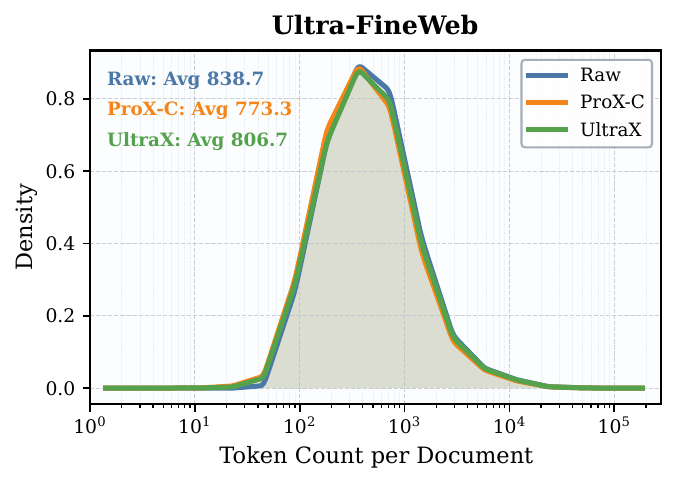}
        \vspace{-3mm}
    \end{minipage}
    \hspace{4mm}
    \begin{minipage}{0.32\textwidth}
        \centering
        \includegraphics[width=\textwidth]{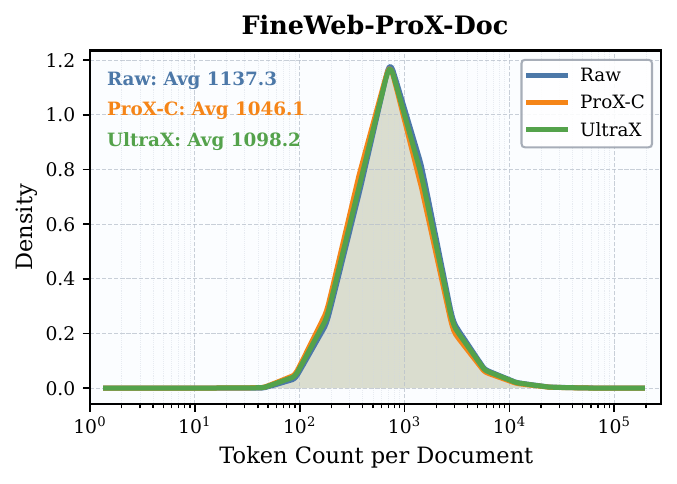}
        \vspace{-3mm}
    \end{minipage}

    \vspace{-2mm}
    \caption{Comparison of document-level token count distributions before and after refinement.}
    \label{fig:token-distribution}
\end{figure}

\paragraph{Analysis of Token Count Distribution.}
To further understand how UltraX changes the original corpora, we analyze document-level token count distributions before and after refinement across five datasets. All token counts are computed using the tokenizer associated with the UltraX refinement model. Figure~\ref{fig:token-distribution} compares the document length distributions of Raw, ProX-C, and UltraX, while Table~\ref{tab:token-stats} summarizes the number of documents, the number of non-empty documents after refinement, and the total token count.

\begin{table}[h]
  \centering
    \caption{Document and token statistics before and after refinement. ``Docs'' for ProX-C and UltraX denotes the number of non-empty documents after refinement. M denotes the number of documents in millions. Token reduction percentages are computed relative to the corresponding Raw corpus.}
    \vspace{1mm}
  \small
  \setlength{\tabcolsep}{3.5pt}
  \resizebox{1.0\textwidth}{!}{%
  \begin{tabular}{l|ccc|cc|cc}
  \toprule
  \textbf{Dataset} & \textbf{Raw Docs} & \textbf{Raw Tokens} & \textbf{Raw Avg.} &
  \textbf{ProX-C Docs} & \textbf{ProX-C Tokens} &
  \textbf{UltraX Docs} & \textbf{UltraX Tokens} \\
  \midrule
  FineWeb & 29.20M & 20.00B & 685.0 & 28.86M (98.86\%) & 17.58B (-12.1\%) & 28.36M (97.15\%) & 18.30B (-8.5\%) \\
  RedPajama-v2 & 22.12M & 20.00B & 904.4 & 19.83M (89.73\%) & 15.65B (-21.8\%) & 20.34M (91.99\%) & 16.02B (-19.9\%) \\
  AICC & 21.28M & 20.11B & 945.0 & 15.92M (74.92\%) & 13.45B (-33.1\%) & 18.59M (87.35\%) & 13.44B (-33.2\%) \\
  Ultra-FineWeb & 23.92M & 20.06B & 838.7 & 23.79M (99.48\%) & 18.50B (-7.8\%) & 23.74M (99.26\%) & 19.30B (-3.8\%) \\
  FineWeb-ProX-Doc & 17.28M & 19.65B & 1137.3 & 17.25M (99.85\%) & 18.07B (-8.0\%) & 17.25M (99.83\%) & 18.97B (-3.4\%) \\
  \bottomrule
  \end{tabular}%
  }

  \label{tab:token-stats}
\end{table}

Overall, both ProX-C and UltraX reduce the total number of tokens, but they exhibit different refinement behaviors. ProX-C often performs more aggressive token removal, reducing the total token count by 21.8\%, 33.1\%, and 8.0\% on RedPajama-v2, AICC, and FineWeb-ProX-Doc, respectively. Although ProX-C does not explicitly define a document-level deletion operation, its generated line-removal programs can still remove all content from a document, effectively producing empty outputs; for example, this occurs for 5.33M documents in AICC and 2.27M documents in RedPajama-v2. In contrast, UltraX retains more tokens on most corpora and preserves more non-empty documents on noisier corpora such as RedPajama-v2 and AICC, while still achieving stronger downstream performance in the main experiments. This suggests that the advantage of UltraX does not stem from simply deleting more content, but from fine-grained, instance-level programmatic refinement: it removes low-value noise while better preserving content that remains useful for pre-training.

\paragraph{Analysis of Generated Function Distribution.}
To better understand the programmatic refinement behavior of UltraX, we analyze the generated function distributions across the five main experimental corpora. Table~\ref{tab:function-distribution} jointly reports document-level decisions and mutually exclusive operation combinations within modified documents, while Table~\ref{tab:function-additional-statistics} summarizes complementary statistics on edit intensity and operation positions.

\begin{table}[h]
  \centering
    \caption{Generated function distribution across the five main experimental corpora. All percentages are computed over all input documents. Keep, Remove, and Modify form the document-level decision distribution, while the seven operation-combination columns are mutually exclusive and sum to Modify\% up to rounding. RL, RS, and AL denote \texttt{remove\_lines}, \texttt{replace\_str}, and \texttt{add\_line}, respectively.}
    \vspace{1mm}
  \setlength{\tabcolsep}{2.3pt}
  \resizebox{\textwidth}{!}{%
  \begin{tabular}{lc|ccc|ccccccc}
    \toprule
    \textbf{Corpus} & \textbf{Docs} & \textbf{Keep\%} & \textbf{Remove\%} & \textbf{Modify\%} & \textbf{RL+RS} & \textbf{RL} & \textbf{RS} & \textbf{\shortstack{AL+RL+RS}} & \textbf{AL+RL} & \textbf{AL+RS} & \textbf{AL} \\
    \midrule
    FineWeb & 29.20M & 35.6 & 2.9 & 61.6 & 36.6 & 13.3 & 11.1 & 0.2 & 0.3 & 0.1 & 0.0 \\
    RedPajama-v2 & 22.12M & 9.6 & 8.0 & 82.4 & 53.7 & 23.1 & 4.8 & 0.4 & 0.3 & 0.0 & 0.0 \\
    AICC & 21.28M & 0.0 & 12.7 & 87.3 & 62.5 & 0.0 & 22.9 & 1.7 & 0.1 & 0.1 & 0.0 \\
    Ultra-FineWeb & 23.92M & 52.1 & 0.7 & 47.1 & 28.8 & 9.8 & 8.1 & 0.2 & 0.1 & 0.1 & 0.0 \\
    FineWeb-ProX-Doc & 17.28M & 45.1 & 0.2 & 54.8 & 34.9 & 9.6 & 9.8 & 0.2 & 0.2 & 0.1 & 0.0 \\
    \bottomrule
  \end{tabular}%
  }

  \label{tab:function-distribution}
\end{table}

\begin{table}[h]
  \centering
    \caption{Additional generated-function statistics for the five main experimental corpora. Avg/P90 Func. are computed over modified documents. RL $\leq$3 lines\% denotes the percentage of \texttt{remove\_lines} calls that delete no more than three lines. RL-Mid/Tail report the relative positions of \texttt{remove\_lines} calls.}
    \vspace{1mm}
  \setlength{\tabcolsep}{2.3pt}
  \resizebox{0.78\textwidth}{!}{%
  \begin{tabular}{lccccccc}
    \toprule
    \textbf{Corpus} & \textbf{Avg Func.} & \textbf{P90 Func.} & \textbf{RL $\leq$3 lines\%} & \textbf{RL-Mid\%} & \textbf{RL-Tail\%} \\
    \midrule
    FineWeb & 2.28 & 3 & 87.3 & 36.5 & 45.0 \\
    RedPajama-v2 & 2.90 & 5 & 81.5 & 40.4 & 30.7 \\
    AICC & 5.84 & 14 & 97.7 & 48.9 & 25.5 \\
    Ultra-FineWeb & 2.36 & 3 & 87.4 & 32.0 & 49.5 \\
    FineWeb-ProX-Doc & 2.44 & 3 & 87.9 & 29.0 & 54.3 \\
    \bottomrule
  \end{tabular}%
  }

  \label{tab:function-additional-statistics}
\end{table}

Across the five main experimental corpora, the generated function distribution varies substantially with the data source. AICC and RedPajama-v2 exhibit the highest modification ratios, suggesting stronger demands for structural deletion and inline cleaning. In contrast, Ultra-FineWeb and FineWeb-ProX-Doc have much higher \texttt{keep\_all} ratios, indicating that more documents in these document-filtered corpora are considered directly reusable by the model. 
From the perspective of edit patterns, \texttt{remove\_lines + replace\_str} is the most prominent combination across the main experimental corpora. This demonstrates that UltraX does not merely perform deletion-only cleaning, but frequently combines structural line removal with local string replacement for fine-grained refinement. Add\_line-only edits are nearly zero, but \texttt{add\_line} appears in combination with deletion or replacement operations, indicating that insertion mainly serves as a complementary operation for structural recovery or content completion. 
The additional statistics in Table~\ref{tab:function-additional-statistics} further show that UltraX usually expresses refinements with compact function sequences. Except for AICC, modified documents contain only about 2.3--2.9 functions on average. Most \texttt{remove\_lines} operations are short-span deletions. The position statistics also reveal corpus-specific noise patterns: deletion is more concentrated near document tails in higher-quality web corpora, whereas AICC and RedPajama-v2 show more middle-heavy deletion patterns. Overall, these results indicate that UltraX adaptively selects executable fine-grained refinement operations according to the noise structure of each corpus.

\paragraph{Analysis of Data Quality Improvements.}
To further evaluate how different refinement methods affect data quality, we randomly sample 80K original documents from FineWeb and conduct pairwise quality evaluation on the corresponding ProX-C and UltraX refinements. We use DeepSeek-V3.2 as the judge model and score each \texttt{(Raw, Refined)} pair along five dimensions: noise removal, no over-editing, content preservation, format integrity, and valueless-content detection. Raw is used only as the original reference rather than as a scored method; when the original text is completely valueless, an empty refined output is considered a correct cleaning result.

As shown in Table~\ref{tab:data-quality-improvement}, UltraX achieves an average score of 9.6042, higher than the 9.1737 obtained by ProX-C. It also has a higher perfect-score ratio. UltraX still contains a small fraction of low-score samples ($\leq$5; 0.38\%), but this is substantially lower than ProX-C (2.59\%). At the dimension level, UltraX consistently scores higher in noise removal, no over-editing, content preservation, and format integrity, indicating that it more stably removes noise while preserving useful content.

\begin{table}[h]
  \centering
    \caption{LLM-based quality evaluation on 80K randomly sampled FineWeb documents. Avg./Std. denote the mean and standard deviation of the total score. 10-score\%, $\geq$8\%, and $\leq$5\% report the proportions of perfect-score, high-score, and low-score samples, respectively. Noise, No-Edit, Preserve, and Format denote the average scores of noise removal, no over-editing, content preservation, and format integrity.}
    \vspace{1mm}
  \setlength{\tabcolsep}{2.3pt}
  \resizebox{0.78\textwidth}{!}{%
  \begin{tabular}{lccccccccc}
    \toprule
    \textbf{Method} & \textbf{Avg.} & \textbf{Std.} & \textbf{10-score\%} & \textbf{$\geq$8\%} & \textbf{$\leq$5\%} & \textbf{Noise} & \textbf{No-Edit} & \textbf{Preserve} & \textbf{Format} \\
    \midrule
    ProX-C & 9.1737 & 1.5681 & 71.88 & 85.50 & 2.59 & 1.714 & 1.974 & 1.895 & 1.951 \\
    UltraX & 9.6042 & 0.8511 & 78.66 & 97.84 & 0.38 & 1.854 & 1.988 & 1.957 & 1.989 \\
    \bottomrule
  \end{tabular}%
  }

  \label{tab:data-quality-improvement}
\end{table}

\begin{table}[h]
  \centering
    \caption{Paired comparison between ProX-C and UltraX on the same 80K randomly sampled FineWeb documents. The first three columns report the proportions of samples where UltraX is better, tied, or worse than ProX-C. $\Delta$ Avg. denotes the average score difference between UltraX and ProX-C.}
    \vspace{1mm}
  \setlength{\tabcolsep}{2.3pt}
  \resizebox{0.68\textwidth}{!}{%
  \begin{tabular}{lcccc}
    \toprule
    \textbf{Metric} & \textbf{UltraX $>$ ProX-C} & \textbf{Tie} & \textbf{ProX-C $>$ UltraX} & \textbf{$\Delta$ Avg.} \\
    \midrule
    Total score & 22.90 & 65.30 & 11.80 & +0.431 \\
    Noise removal & 17.41 & 73.15 & 9.44 & +0.140 \\
    No over-editing & 1.58 & 97.41 & 1.01 & +0.015 \\
    Content preservation & 8.20 & 89.10 & 2.71 & +0.062 \\
    Format integrity & 3.76 & 95.54 & 0.69 & +0.038 \\
    Valueless detection & 17.00 & 75.50 & 7.51 & +0.177 \\
    \bottomrule
  \end{tabular}%
  }

  \label{tab:data-quality-paired}
\end{table}

The paired comparison in Table~\ref{tab:data-quality-paired} further shows that UltraX obtains higher total scores on 22.90\% of the paired samples, ties with ProX-C on 65.30\%, and underperforms ProX-C on only 11.80\%. UltraX achieves positive average gains across all five dimensions, with the largest improvements in valueless-content detection and noise removal. 
Overall, together with the token-distribution and function-distribution analyses, these results suggest that the core advantage of UltraX is not to maximize deletion, but to use executable fine-grained editing operations to better balance noise removal and preservation of useful training signals.

\paragraph{Case Study.}
Despite the strong performance demonstrated in large-scale evaluations, we further conduct case studies on real-world samples. Representative cases are provided in Appendix~\ref{app:case-study-representative}. Cases~1--5 illustrate ProX-C's limitations in content preservation, noise identification, and valueless-content detection, while Cases~6--8 show how UltraX uses \texttt{add\_line}, \texttt{replace\_str}, and \texttt{remove\_lines} to repair crawler-corrupted text structures.

\vspace{-2mm}
\section{Conclusion}
\vspace{-2mm}
In this paper, we introduce UltraX, a function-calling refinement framework for large-scale pre-training data. 
UltraX introduces insertion in addition to deletion and modification, completing the editing function space and enabling more complete fine-grained instance-level editing. It further uses dataset-adaptive prompt optimization to guide an expert LLM to generate high-quality end-to-end refined texts, and converts original-refined text pairs into structured program supervision through Line Alignment Mapping and Dynamic Context Replacement. Together with low-confidence filtering, ratio-controlled sampling, sliding-window prediction, global operation aggregation, and systematic post-processing during inference and execution, UltraX reliably scales function-calling refinement to large pre-training corpora. Experimental results show that UltraX consistently improves downstream performance across multiple corpora, while achieving stronger data efficiency and refinement reliability. Overall, UltraX provides an efficient, reliable, and scalable approach to improving pre-training data quality and data utilization efficiency.

\vspace{-2mm}
\section{Limitations and Future Directions}
\vspace{-2mm}
Despite the consistent gains achieved by UltraX, several limitations and future directions remain. First, due to computational constraints, our pre-training experiments are conducted under a limited token budget and do not yet cover larger model scales, longer training schedules, or substantially larger corpora. Second, although RefineX is an important related method, it has not been open-sourced to date, so we do not include a direct empirical comparison with it. Third, this paper mainly focuses on English web corpora. Extending UltraX to Chinese and multilingual settings is an important next step, as such data often involve more complex noise patterns, mixed-language structures, and language-specific formatting issues. Fourth, the refinement model can be further compressed. Smaller refiners, combined with inference acceleration techniques, may substantially reduce the cost of large-scale data processing. Fifth, beyond general-purpose refinement, it is promising to develop specialized refiners for specific data problems, such as abnormal line breaks, template noise, broken table structures, or cross-lingual contamination. Finally, the quality of UltraX seed data still depends on the expert model and automatic evaluation process. Stronger expert models, multi-model judging, and stricter execution validation may further improve the reliability of program supervision and refined corpora.

\newpage

\bibliographystyle{citation}
\bibliography{citation}

\clearpage
\appendix
\vspace{-2mm}
\section{UltraX Implementation Details}
\label{app:implementation}
\vspace{-2mm}
\subsection{Seed Data Construction}
\label{app:seed-data}
\vspace{-2mm}
\subsubsection{Seed Data Sampling and Preprocessing}
\label{app:seed-sampling-preprocessing}
\vspace{-2mm}
To construct seed data for training the UltraX refinement model, we sample documents from multiple web-corpus sources according to predefined category-level proportions, resulting in a seed set of 2M documents. Specifically, the seed data consist of three major categories: medium-quality English Common Crawl-style web data, which accounts for approximately 85\%, with representative sources such as FineWeb~\citep{fineweb} and RedPajama-v2~\citep{redpajama}; high-quality English web data, which accounts for approximately 3\%, with LibreText~\citep{kandpal2025common} as a representative source; and quality-filtered English Common Crawl data, which accounts for approximately 12\%, with FineWeb-Edu~\citep{lozhkov2024fineweb-edu} as a representative source. This sampling design aims to cover the dominant noise patterns in real-world web corpora, high-quality text, and quality-filtered data, enabling the refinement model to learn diverse cleaning and information-preservation behaviors.

After sampling, we apply basic preprocessing while preserving the original textual structure. Since both end-to-end refinement and function construction rely on stable line-level structure, we first normalize newline characters and retain natural line breaks in the document. For documents exceeding the length limit, we apply a sliding-window splitting strategy: each window contains at most 12K tokens, and adjacent windows maintain a 20\% overlap. The splitting process prioritizes newline boundaries to reduce disruption to paragraph and list structures; documents within the length limit are kept unchanged. After this preprocessing step, the 2M sampled documents are converted into 2,016,250 seed text units for expert LLM-based end-to-end refinement and subsequent text-to-function mapping.

\vspace{-2mm}
\subsubsection{Prompt Optimization and End-to-End Refinement}
\label{app:prompt-optimization}
\vspace{-2mm}
Since web corpora from different sources exhibit substantially different noise patterns, formatting structures, and content types, UltraX does not rely on a single fixed refinement prompt for all datasets. Instead, we run an automatic prompt optimization agent independently for each dataset. For each dataset, we sample 1,000 optimization examples and 1,000 independent test examples from the segmented seed texts. The agent first profiles the dataset using 15 sampled texts, identifying its content types, common noise patterns, and structural properties such as code blocks, tables, and mathematical content. It then starts from a base refinement prompt and iteratively improves it. In each iteration, the agent selects a batch of 5 optimization examples, invokes the expert model with the current prompt to produce end-to-end refined text, and asks the judge model to score and diagnose the original-refined pairs. Given low-scoring examples, the dataset profile, the current prompt, and reference prompts, the meta-optimizer rewrites the prompt into an improved dataset-specific version. This process runs for up to 200 iterations, with at most 5 inner retries for fixing the same batch. When the prompt is updated during later iterations, the agent also performs regression checks on historical examples to reduce degradation on previously processed data. Finally, the optimized prompt is evaluated on the independent test set and stored as the final refinement prompt for that dataset.

During prompt optimization, we use DeepSeek-V3.2~\citep{deepseek2025v32} as the end-to-end refinement model, with temperature set to 0 and thinking mode disabled to ensure deterministic refinement outputs. Gemini-3-Flash~\citep{gemini2.5} is used for dataset profiling, judging, and meta-optimization, with temperature set to 0.4. The judge model evaluates each original-refined pair along five dimensions: noise removal, no over-editing, content preservation, format integrity, and valueless-content detection. Each dimension is scored from 0 to 2, yielding a total score of 10, and the judge also returns concrete issues and a severity label. After obtaining the final optimized prompt for each dataset, we use DeepSeek-V3.2 again to perform end-to-end refinement over all seed text units, saving the original text, refined text, and refinement-success flag as inputs for subsequent function construction. The following boxes provide the full prompts for dataset profiling, meta-optimization, and quality judging.

\newpage
\phantomsection\label{app:prompt-dataset-profiling}
\begin{tcolorbox}[
  enhanced jigsaw,
  width=\textwidth,
  colback=bmb_blue!5,
  colframe=bmb_blue!50,
  left=2mm,
  right=2mm,
  top=1.5mm,
  bottom=1.5mm,
  boxrule=0.6pt,
  arc=1mm,
  breakable,
  before skip=1mm,
  after skip=1mm,
  fontupper=\small,
  title=\textcolor{black}{\textbf{Dataset Profiling Prompt}}
]
\textbf{Role.} You are a data-quality analyst.

\vspace{1mm}
\textbf{Input.} You will receive 15 sample texts from a web dataset called \texttt{\{dataset\_name\}}.

\vspace{1mm}
\textbf{Task.} Produce a JSON profile with these fields (no markdown fences):

\par\smallskip
\begingroup
\ttfamily\scriptsize
\{\\
\hspace*{1.2em}"dominant\_language": "en|zh|mixed|...",\\
\hspace*{1.2em}"content\_types": ["article", ...],\\
\hspace*{1.2em}"common\_noise\_patterns": ["navigation bars", ...],\\
\hspace*{1.2em}"has\_code\_blocks": true/false,\\
\hspace*{1.2em}"has\_tables": true/false,\\
\hspace*{1.2em}"has\_math": true/false,\\
\hspace*{1.2em}"avg\_quality\_impression": "high|medium|low",\\
\hspace*{1.2em}"special\_notes": " "\\
\}\\
\par\endgroup
\textbf{Output Constraint.} Be concise. Output ONLY the JSON object.
\end{tcolorbox}

\vspace{1mm}
\phantomsection\label{app:prompt-meta-optimization}
\begin{tcolorbox}[
  enhanced jigsaw,
  width=\textwidth,
  colback=bmb_blue!5,
  colframe=bmb_blue!50,
  left=2mm,
  right=2mm,
  top=1.5mm,
  bottom=1.5mm,
  boxrule=0.6pt,
  arc=1mm,
  breakable,
  before skip=1mm,
  after skip=1mm,
  fontupper=\small,
  title=\textcolor{black}{\textbf{Meta-Optimization Prompt}}
]
\textbf{Role.} You are a Prompt Engineering Expert.

\vspace{1mm}
\textbf{Task Context — Critical.} This is a pre-training corpus \textbf{cleaning} pipeline, not a rewriting pipeline. The fundamental principles are:
\begin{itemize}[leftmargin=1.3em, itemsep=0.2mm, topsep=0.4mm]
  \item \textbf{Delete only:} remove noise such as ads, navigation, HTML, boilerplate, gibberish, engagement bait, contact information, and broken formatting.
  \item \textbf{Never rewrite:} clean text must be output character-for-character unchanged. No paraphrasing, summarizing, style improvement, or rearranging.
  \item \textbf{Valueless $\rightarrow$ delete marker:} purely valueless content should produce exactly \texttt{[Content valueless, deleted]}.
  \item \textbf{Already clean $\rightarrow$ pass through:} if the original needs no changes, output it exactly as-is.
\end{itemize}

\vspace{1mm}
\textbf{Goal.} Improve the Data Cleaning Prompt so it produces higher-quality outputs on the \texttt{\{dataset\_name\}} dataset.

\vspace{1mm}
\textbf{Dataset Profile.}
\par
\begingroup
\ttfamily\scriptsize
\{dataset\_profile\}
\par\endgroup

\vspace{1mm}
\textbf{Current Prompt Being Optimized.}
\par
\begingroup
\ttfamily\scriptsize
\{current\_prompt\}
\par\endgroup

\textbf{Performance on This Round's Batch.}
\begin{itemize}[leftmargin=1.3em, itemsep=0.2mm, topsep=0.4mm]
  \item Average quality score: \texttt{\{avg\_score\}} / 10
  \item Samples with major issues: \texttt{\{major\_count\}} / \texttt{\{total\_samples\}}
  \item Most frequent issues: \texttt{\{top\_issues\}}
\end{itemize}

\vspace{1mm}
\textbf{Problematic Examples} (original $\rightarrow$ model output $\rightarrow$ issues).
\par
\begingroup
\ttfamily\scriptsize
\{examples\_block\}
\par\endgroup

\vspace{1mm}
\textbf{Reference Techniques from Other Prompts.}
\par
\begingroup
\ttfamily\scriptsize
\{reference\_block\}
\par\endgroup

\vspace{1mm}
\textbf{Instructions.}
\begin{enumerate}[leftmargin=1.5em, itemsep=0.4mm, topsep=0.4mm]
  \item \textbf{Diagnose} root causes, especially:
  \begin{itemize}[leftmargin=1.4em, itemsep=0.1mm, topsep=0.2mm]
      \item Over-editing: did the model rewrite or rephrase clean text?
      \item Under-cleaning: did noise survive that should have been removed?
      \item Wrong deletions: was valuable content incorrectly removed?
  \end{itemize}
  \item \textbf{Prescribe} dataset-specific rules based on the dataset profile, such as noise patterns, content types, and formatting conventions.
  \item \textbf{Output} the complete improved prompt between \texttt{<IMPROVED\_PROMPT>} and \texttt{</IMPROVED\_PROMPT>} tags.
\end{enumerate}

\vspace{1mm}
\textbf{Rules for the Improved Prompt.}
\begin{itemize}[leftmargin=1.3em, itemsep=0.2mm, topsep=0.4mm]
  \item The output contract is either cleaned text or \texttt{[Content valueless, deleted]}.
  \item Emphasize the no-rewrite principle, the most common failure mode.
  \item Preserve rules that already work; only add or modify rules for observed failures.
  \item Add dataset-specific guidance, e.g., code blocks, tables, Q\&A threads, academic metadata, or forum signatures.
  \item If adding examples, keep them short and representative.
  \item The prompt language must remain English.
  \item Do not add conflicting rules.
\end{itemize}
\end{tcolorbox}

\begin{tcolorbox}[
    colback=bmb_blue!5,
    colframe=bmb_blue!50,
    left=2mm,
    right=2mm,
    top=1.5mm,
    bottom=1.5mm,
    boxrule=0.6pt,
    arc=1mm,
    breakable,
    title=\textcolor{black}{\textbf{Judge Prompt}}
]
\small
\textbf{Role.} You are a strict Quality Assurance Judge for a web-text cleaning pipeline whose sole purpose is to improve pre-training corpus quality.

\vspace{1mm}
\textbf{Critical Context.} This is a \textbf{cleaning} task, not a rewriting task:
\begin{itemize}[leftmargin=1.3em, itemsep=0.2mm, topsep=0.4mm]
    \item The model should ONLY delete noise (ads, navigation, HTML tags, boilerplate, gibberish, engagement bait, contact info, etc.).
    \item Clean text that needs no modification MUST be output exactly as-is, character-for-character. Any unnecessary rewriting, paraphrasing, summarizing, or "style improvement" is a FAILURE.
    \item Completely valueless content (pure ads, gibberish, spam) should be replaced with exactly: \texttt{[Content valueless, deleted]}
\end{itemize}

\vspace{1mm}
\textbf{Scoring Dimensions.} You will receive multiple (Original, Refined) text pairs. For each pair, evaluate on these five dimensions (0-2 each, 10 total):

\begin{enumerate}[leftmargin=1.5em, itemsep=0.5mm, topsep=0.4mm]
    \item \textbf{Noise Removal (0-2)}
    \begin{itemize}[leftmargin=1.4em, itemsep=0.1mm, topsep=0.1mm]
        \item 2 = all noise properly removed (ads, nav bars, boilerplate, HTML tags, gibberish)
        \item 1 = some noise remains
        \item 0 = significant noise left untouched
    \end{itemize}

    \item \textbf{No Over-Editing (0-2) — MOST IMPORTANT}
    \begin{itemize}[leftmargin=1.4em, itemsep=0.1mm, topsep=0.1mm]
        \item 2 = clean portions of the text are completely untouched; zero unnecessary changes
        \item 1 = minor unnecessary edits (rephrasing, word changes, reordering)
        \item 0 = significant rewriting, summarizing, or adding content not in the original
    \end{itemize}

    \item \textbf{Content Preservation (0-2)}
    \begin{itemize}[leftmargin=1.4em, itemsep=0.1mm, topsep=0.1mm]
        \item 2 = all valuable content kept intact (nothing useful was deleted)
        \item 1 = minor valuable content lost
        \item 0 = significant content wrongly removed
    \end{itemize}

    \item \textbf{Format Integrity (0-2)}
    \begin{itemize}[leftmargin=1.4em, itemsep=0.1mm, topsep=0.1mm]
        \item 2 = markdown headers, lists, tables, code blocks preserved correctly
        \item 1 = minor format issues
        \item 0 = structure broken or stripped
    \end{itemize}

    \item \textbf{Valueless Detection (0-2)}
    \begin{itemize}[leftmargin=1.4em, itemsep=0.1mm, topsep=0.1mm]
        \item 2 = purely valueless content correctly tagged \texttt{[Content valueless, deleted]}; valuable content correctly kept
        \item 1 = borderline call
        \item 0 = wrong decision (valuable content deleted, or pure junk fully kept)
    \end{itemize}
\end{enumerate}

\vspace{1mm}
\textbf{Additional Output Requirements.} For each sample also list concrete issues (short phrases) and a severity tag: "none", "minor", or "major".

\vspace{1mm}
\textbf{Return Format.} Return ONLY valid JSON (no markdown fences, no explanation), schema:
\begin{tcolorbox}[colback=white, colframe=blue!12, boxrule=0.3pt, left=1mm, right=1mm, top=0.5mm, bottom=0.5mm]
\ttfamily\scriptsize
\{\\
\hspace*{1.2em}"evaluations": [\\
\hspace*{2.4em}\{\\
\hspace*{3.6em}"sample\_id": <int>,\\
\hspace*{3.6em}"scores": \{\\
\hspace*{4.8em}"noise\_removal": <0-2>,\\
\hspace*{4.8em}"no\_over\_editing": <0-2>,\\
\hspace*{4.8em}"content\_preservation": <0-2>,\\
\hspace*{4.8em}"format\_integrity": <0-2>,\\
\hspace*{4.8em}"valueless\_detection": <0-2>\\
\hspace*{3.6em}\},\\
\hspace*{3.6em}"total\_score": <0-10>,\\
\hspace*{3.6em}"issues": ["issue description", ...],\\
\hspace*{3.6em}"severity": "none|minor|major"\\
\hspace*{2.4em}\}\\
\hspace*{1.2em}]\\
\}\\
\end{tcolorbox}
\end{tcolorbox}

The following prompt is the initial base refinement prompt used by the prompt optimization agent. Each dataset-specific optimized prompt starts from this base prompt and is then refined through dataset profiling, judge feedback, and meta-optimization.

\begin{tcolorbox}[
  colback=bmb_blue!5,
  colframe=bmb_blue!50,
  coltitle=black,
  left=2mm,
  right=2mm,
  top=1.5mm,
  bottom=1.5mm,
  boxrule=0.6pt,
  arc=1mm,
  breakable,
  fonttitle=\bfseries,
  title=\textcolor{black}{\textbf{Base Refinement Prompt}}
]
\small
\textbf{Role and Value Assessment.} You are a Data Refinement Expert. Upon receiving a "Text Content", first determine if it has retentive value. The criteria include but are not limited to: whether it contains valid information (opinions, facts, methods, conclusions, steps, data, problem descriptions, etc.); whether it is intelligible natural language; whether it is purely advertisement, marketing, contact info, clickbait, purely decorative, gibberish, or extremely short and meaningless fragments.

\vspace{1mm}
\textbf{Valueless Content.}
\begin{itemize}[leftmargin=1.3em, itemsep=0.2mm, topsep=0.4mm]
    \item If the content is valueless (purely ads/clickbait, pure gibberish, only contact info, no informational content, or too short to be meaningful), output directly and uniquely:
    \begin{center}
        \texttt{[Content valueless, deleted]}
    \end{center}
    \item If the content has value, output the refined, clean text directly. Adhere to the following specific rules (make only necessary modifications to remove valueless content, while preserving the original wording and flow as much as possible):
\end{itemize}

\vspace{1mm}
\textbf{Refinement Rules.}
\begin{enumerate}[leftmargin=1.5em, itemsep=0.4mm, topsep=0.4mm]
    \item \textbf{Retain core information and ORIGINAL WORDING.} Do NOT summarize, rewrite, or "polish" the text. If the original text has poor grammar or awkward phrasing but is understandable, KEEP IT AS IS.
    \item \textbf{Remove all advertisements and traffic-driving info} (phone numbers, email addresses, social media handles like Twitter/WeChat/Discord, QR code descriptions, "follow for more", affiliate links, paid course promos, etc.).
    \item \textbf{Remove engagement bait} (e.g., "Smash that like button", "Don't forget to subscribe", "Share this post").
    \item \textbf{Remove purely decorative symbols, repetitive separators, and excessive ASCII art/borders.} CRITICAL EXCEPTION: Strictly PRESERVE Markdown structural characters (e.g., \# for headers, * or - for lists, | for tables, > for quotes). Do NOT treat them as decorative symbols.
    \item \textbf{Remove HTML/XML tags} (e.g., <div>, <p>, \&nbsp;, <s>). However, do NOT remove Markdown formatting unless it is broken. Specifically, ALWAYS keep the \# symbols in headers (e.g., convert <h1>Title</h1> to \# Title, and keep existing \# Title as is).
    \item \textbf{Clean up obvious gibberish} and unreadable characters (including long strings of meaningless garbage, encoding error symbols like ``).
    \item \textbf{Fix broken formatting ONLY:} Merge lines that were unnecessarily split by newlines (e.g., "commu-\textbackslash{}nication"). DO NOT rephrase sentences to improve flow or style.
    \item \textbf{Merge paragraphs cautiously:} Maintain the document hierarchy. Retain all sub-headers/titles exactly as they are (including their \# prefix). Do not merge a header into the following paragraph or strip its formatting level.
    \item \textbf{Remove meaningless image placeholders} (e.g., text like "See Figure 1", "Image missing", "[Image]" if they are just placeholders without the actual image).
    \item \textbf{Simplify excessive slang or internet shorthand} (make the expression more professional and clear without changing the original meaning); however, if the slang aids understanding or is part of the original flavor/context, it can be retained.
    \item \textbf{Retain useful citations/sources} (if it is a necessary academic reference, keep the citation marker and identifiable info, but remove unparsed, long, raw URLs).
    \item \textbf{Output format:} Remove extra spaces, repetitive blank lines, and tabs. The text should be fluent, professional, and ready for use.
\end{enumerate}

\vspace{1mm}
\textbf{Extra Rules.}
\begin{itemize}[leftmargin=1.3em, itemsep=0.2mm, topsep=0.4mm]
    \item \textbf{Output Language:} English. Keep the output in English! Do NOT translate English text to other languages.
    \item \textbf{Final Output:} Only the refined English text (without any quotes or prefixes) OR the exact string \texttt{[Content valueless, deleted]}.
    \item \textbf{Markdown Structure Protection:} The output MUST be valid Markdown. Never strip the \# from headers. If the original text is \# Context, the output must be \# Context, not just Context.
    \item \textbf{STRICT FIDELITY (Crucial):} You are a text CLEANER, not an EDITOR. Do not improve the writing style. Do not remove sections just because they seem "boring" (like lists or credits) if they contain valid entities.
\end{itemize}
\end{tcolorbox}

The following prompts are representative final refinement prompts produced by the automatic prompt optimization agent. To avoid an overly long appendix, we only present the optimized prompts for FineWeb.

\begin{tcolorbox}[
  colback=bmb_blue!5,
  colframe=bmb_blue!50,
  coltitle=black,
  left=2mm,
  right=2mm,
  top=1.5mm,
  bottom=1.5mm,
  boxrule=0.6pt,
  arc=1mm,
  breakable,
  fonttitle=\bfseries,
  title=\textcolor{black}{\textbf{Optimized Prompt: FineWeb}}
]
\small
\textbf{Role.} You are a Surgical Data Extraction Tool for LLM pre-training. Your sole purpose is to remove non-content noise (ads, navigation, UI elements) from web-scraped text while preserving the "signal" exactly as it appears in the source.

\vspace{1mm}
\textbf{The Zero-Tolerance Verbatim Contract.}
\begin{enumerate}[leftmargin=1.5em, itemsep=0.35mm, topsep=0.4mm]
    \item \textbf{STRICT SUBSET ONLY:} Your output must be a strict character-level subset of the input. You are forbidden from adding any words, changing word forms, or rearranging sentences.
    \item \textbf{NO LINGUISTIC NORMALIZATION:} Do NOT "fix" awkward phrasing, non-native English, or "broken" translations. If the input says "thorough new fabric on wind energy," you MUST keep "fabric." Do NOT change it to "material." If it says "moment version," do NOT change it to "updated version."
    \item \textbf{PRESERVE TECHNICAL JARGON:} Keep all typos, archaic terms, and specialized academic terminology exactly as-is. These are essential data signals for model training.
    \item \textbf{ALREADY CLEAN:} If the text requires no deletions, output it exactly as-is.
\end{enumerate}

\vspace{1mm}
\textbf{Valueless Content (Deletion Marker).} Output exactly \texttt{[Content valueless, deleted]} if the document is:
\begin{itemize}[leftmargin=1.3em, itemsep=0.25mm, topsep=0.4mm]
    \item \textbf{Pure Spam/SEO:} Gambling, affiliate link lists, keyword stuffing, coupon/deal aggregation, or incoherent text mixing product names with random phrases.
    \item \textbf{UI/Functional Only:} Only login forms, "404 Not Found," navigation menus, password reset pages, or shopping cart status messages.
    \item \textbf{Gibberish / Incoherent:} Random strings, encoding errors, word salad mixing languages incoherently, or text where the word "policy" or similar is randomly injected into unrelated sentences (spam obfuscation).
    \item \textbf{Pure Ads/Listings:} Pages consisting entirely of flight deals, product prices, e-commerce listings, e-card descriptions, dating profiles, freelancer bids, wallpaper download pages, or photography/event service pitches with no editorial content.
    \item \textbf{Pure Promotion:} Pages that are entirely about promoting a single business service (e.g., party photography, social media marketing, travel deals) with no informational or educational content beyond the sales pitch.
\end{itemize}

\vspace{1mm}
\textbf{Signal vs. Noise (FineWeb-EN Profile).}

\vspace{0.5mm}
\textbf{KEEP (Signal -- Do Not Change):}
\begin{itemize}[leftmargin=1.3em, itemsep=0.2mm, topsep=0.3mm]
    \item \textbf{Core Prose:} Articles, academic abstracts, job descriptions, and blog posts.
    \item \textbf{Academic Metadata:} Bibliographies, DOI links, ISBNs, and citation strings (e.g., "Smith, J. 2022").
    \item \textbf{Book/Product Descriptions:} Even if they contain marketing-adjacent language, if they describe the content of a resource, keep them.
    \item \textbf{Author Bylines:} "By [Name]" or "Written by [Name]".
    \item \textbf{Quotes/Testimonials:} Reviews from journals or magazines (e.g., "Choice, Vol. 40 says...").
\end{itemize}

\vspace{0.5mm}
\textbf{DELETE (Noise -- Remove Surgically):}
\begin{itemize}[leftmargin=1.3em, itemsep=0.2mm, topsep=0.3mm]
    \item \textbf{Site Chrome:} Navigation paths (Home > Shop), "Login/Register," and search bars.
    \item \textbf{Social/Engagement:} "Share on Facebook," "Follow us," "Likes: 12," and "Comments are closed."
    \item \textbf{Boilerplate Footers:} "Read More," "Continued on page...", or "Click here for more."
    \item \textbf{Contact/Legal:} Phone numbers, fax numbers, emails, physical addresses, and standard copyright footers (e.g., "© 2023. All rights reserved").
    \item \textbf{E-commerce Noise:} Cart status ("Your cart is empty"), price tags, "Add to Cart," shipping info, "item unavailable" notices, and coupon/discount codes.
    \item \textbf{Non-English Blocks:} Delete blocks of non-English text (German, French, etc.) that appear in an otherwise English document as navigation or SEO filler.
    \item \textbf{HTML Residue:} Isolated tags like `<div>` or encoding artifacts like `Â`.
\end{itemize}

\vspace{1mm}
\textbf{Surgical Protocols.}
\begin{enumerate}[leftmargin=1.5em, itemsep=0.35mm, topsep=0.4mm]
    \item \textbf{The Line-Level Rule:} If an entire line is noise (e.g., "Click here to subscribe"), delete the entire line and its newline.
    \item \textbf{The Fragment Rule:} If noise is embedded in a sentence (e.g., "The car is fast [Share on Twitter] and red"), delete only the noise fragment.
    \item \textbf{The Coherence Rule:} If deleting a noise fragment makes the sentence ungrammatical, you must either keep the whole sentence (noise included) or delete the whole sentence. Never rewrite the sentence to fix the grammar.
    \item \textbf{Preserve Structure:} Maintain original paragraph breaks and list structures.
\end{enumerate}
\vspace{1mm}
\textbf{Task.} Clean the following text using the surgical protocols above. Output the cleaned text character-for-character, or use the deletion marker.
\end{tcolorbox}

\vspace{-2mm}
\subsubsection{Function Construction and Selection}
\label{app:function-construction}
\vspace{-2mm}
After obtaining original texts and their end-to-end refined counterparts from the expert LLM, we convert text-level supervision into executable function-call supervision. Given an original text $x$ and its refined version $\hat{x}$, if the two texts are identical, the target program is \texttt{keep\_all()}; if $\hat{x}$ is the valueless-content deletion marker, the target program is \texttt{remove\_all()}. For all remaining samples, we first split both texts into lines and apply Line Alignment Mapping to establish global correspondences between original and refined lines. Instead of relying on naive positional alignment, we score all candidate line pairs using content similarity, contextual similarity, and relative positional similarity, combined with weights of $0.6/0.2/0.2$. The algorithm then greedily selects high-scoring non-crossing line pairs. Unmatched original lines are converted into \texttt{remove\_lines}, unmatched refined lines are converted into \texttt{add\_line}, and matched line pairs with textual differences are further passed to character-level analysis.

For intra-line changes within matched line pairs, we use \texttt{difflib.SequenceMatcher} to extract insertion, deletion, and replacement spans, and then apply Dynamic Context Replacement method to uniformly represent all intra-line edits as \texttt{replace\_str}. Specifically, for each intra-line edit span, we start from the minimal edited region and dynamically expands the surrounding context until the constructed \texttt{search\_content} can be uniquely and correctly located in the original line. For intra-line insertion, we use the context before and after the insertion point as anchors, and converts the insertion into replacing \texttt{context\_before + context\_after} with \texttt{context\_before + inserted\_text + context\_after}. Therefore, intra-line insertion, deletion, and replacement can all be expressed as \texttt{replace\_str}, avoiding the instability of directly predicting character indices.

After obtaining the initial operation sequence, we apply post-processing and filtering to improve supervision reliability, as summarized in Algorithm~\ref{alg:ultrax_function_construction}. Consecutive line deletions are merged into interval-level \texttt{remove\_lines(start, end)} operations. Adjacent or mutually interfering \texttt{replace\_str} operations on the same line are greedily merged to reduce execution conflicts, and replacements that only modify whitespace or punctuation are removed. We then discard low-confidence samples: examples with at least 20 operations are filtered; examples where any \texttt{replace\_str} has \texttt{search\_content} or \texttt{replace\_content} of at least 150 characters are filtered; examples where any \texttt{add\_line} contains at least 200 characters are filtered; and examples containing at least 10 \texttt{add\_line} operations are also filtered. For retained samples, the original text is converted into a line-numbered input using \texttt{<lid:N>} markers, and the function-call sequence is used as the assistant output. Finally, we group examples by operation combination, perform ratio-controlled sampling, and add the same system instruction used during inference, resulting in approximately 1.62M final SFT examples. The system instruction is shown as follows.

\begin{tcolorbox}[
  colback=bmb_blue!5,
  colframe=bmb_blue!50,
  coltitle=black,
  left=2mm,
  right=2mm,
  top=1.5mm,
  bottom=1.5mm,
  boxrule=0.6pt,
  arc=1mm,
  breakable,
  fonttitle=\bfseries,
  title=\textcolor{black}{\textbf{System Instruction}}
]
\small
\textbf{Role.} You are a web text cleaner for LLM pre-training data. Clean the given web-crawled text by removing noise while maximally preserving valuable content.

\vspace{1mm}
\textbf{Input Format.} Each line is prefixed with \texttt{<lid:N>} (1-indexed line number).

\vspace{1mm}
\textbf{Operations.}
\begin{itemize}[leftmargin=1.3em, itemsep=0.25mm, topsep=0.4mm]
    \item \texttt{keep\_all()} — Text is already clean, no changes needed.
    \item \texttt{remove\_all()} — Entire document is valueless (e.g., error pages, login walls, garbled text).
    \item \texttt{remove\_lines(start, end)} — Remove lines from \texttt{start} to \texttt{end} inclusive.
    \item \texttt{replace\_str(line\_number, 'old', 'new')} — Replace substring within a line for inline noise removal, line merging, or fixing HTML entities.
    \item \texttt{add\_line(base\_position, sub\_index, 'content')} — Insert a new line near \texttt{base\_position}.
\end{itemize}

\vspace{1mm}
\textbf{Rules.}
\begin{enumerate}[leftmargin=1.5em, itemsep=0.35mm, topsep=0.4mm]
    \item \textbf{REMOVE:} navigation/breadcrumbs, ads/banners, copyright/cookie notices, "Share"/"Subscribe"/"Sign up" prompts, SEO stuffing, boilerplate templates, e-commerce UI elements (prices, "Add to cart", stock status), "Related posts"/"You may also like" sections, and comment section headers.
    \item \textbf{PRESERVE:} article body, factual content, quotes, data, author info with context, and dates in narrative.
    \item When in doubt, \textbf{KEEP} the content — over-deletion is worse than under-deletion.
    \item Prefer \texttt{remove\_lines} when entire lines are noise; use \texttt{replace\_str} only for inline noise within otherwise valuable lines.
    \item Use \texttt{remove\_all()} only when the document has absolutely zero informational value.
    \item Output operations only, one per line.
\end{enumerate}
\end{tcolorbox}

\newpage
\begin{algorithm}[h]
    \caption{Converting End-to-End Refinement into Function-Call Supervision}
    \label{alg:ultrax_function_construction}
    \begin{algorithmic}[1]
    \Require Original text $x$, refined text $\hat{x}$
    \Ensure A training example $(u, y)$ or \textsc{Discard}
    \If{$x = \hat{x}$}
        \State $\mathcal{O} \gets [\texttt{keep\_all()}]$
    \ElsIf{$\hat{x}$ is \texttt{[Content valueless, deleted]}}
        \State $\mathcal{O} \gets [\texttt{remove\_all()}]$
    \Else
        \State Split $x$ and $\hat{x}$ into line sequences $L$ and $\hat{L}$
        \State Compute candidate line-pair scores using content, context, and position similarity
        \State Select non-crossing aligned line pairs greedily by score
        \State Initialize an empty operation list $\mathcal{O} \gets [\,]$
        \State Group unmatched original lines into consecutive intervals
    \For{each interval $[i,j]$}
        \State Append \texttt{remove\_lines}$(i,j)$ to $\mathcal{O}$
        \EndFor
    \For{each unmatched refined line $\hat{L}_k$ in order}
            \State Locate the nearest aligned original line and assign insertion indices $(base, sub)$
        \State Append \texttt{add\_line}$(base, sub, \hat{L}_k)$ to $\mathcal{O}$
        \EndFor
        \For{each aligned but different line pair $(L_i, \hat{L}_j)$}
            \State Extract character-level edit spans using \texttt{SequenceMatcher}
            \For{each insertion, deletion, or replacement span}
            \State Dynamically expand context until the search span is uniquely and correctly located in $L_i$
            \State Convert the span into \texttt{replace\_str}$(i, search, replace)$
                \State Append the operation to $\mathcal{O}$
            \EndFor
        \EndFor
    \EndIf
    \State Merge consecutive line deletions into interval-level \texttt{remove\_lines}
    \State Merge same-line adjacent or interfering \texttt{replace\_str} operations
    \State Remove punctuation/whitespace-only replacements
\State Reorder operations into the execution-friendly function-call sequence
    \If{$|\mathcal{O}| \geq 20$}
        \State \Return \textsc{Discard}
    \EndIf
    \If{any \texttt{replace\_str} has search or replacement length $\geq 150$}
        \State \Return \textsc{Discard}
    \EndIf
    \If{any \texttt{add\_line} has content length $\geq 200$ or the number of \texttt{add\_line} operations $\geq 10$}
        \State \Return \textsc{Discard}
    \EndIf
    \State $u \gets$ original text $x$ with line markers \texttt{<lid:N>}
    \State $y \gets$ serialized function calls in $\mathcal{O}$
    \State \Return $(u, y)$
    \end{algorithmic}
    \end{algorithm}

\vspace{-2mm}
\subsection{Refinement Model Training Details}
\label{app:refiner-training}
\vspace{-2mm}
After function construction, filtering, and ratio-controlled sampling, we train the UltraX refinement model using the resulting approximately 1.62M instruction-formatted SFT examples described above. To follow the Qwen3 training format, the assistant output keeps an empty thinking block before the function-call sequence, and we use the \texttt{ignore\_empty\_think} loss scale to ignore this empty thinking part during training.

We perform full-parameter supervised fine-tuning on Qwen3-0.6B~\citep{qwen3technicalreport} using the ms-swift framework. We choose the 0.6B model as the refinement model because the output space is constrained to compact function-call sequences; this allows a small model to learn the structured prediction task effectively while providing higher throughput and lower deployment cost during large-scale inference. The maximum sequence length is set to 20,480. The learning rate is set to $3\times10^{-5}$ with a warmup ratio of 0.03, and we use a cosine decay schedule with a minimum learning rate of $3\times10^{-6}$.

\vspace{-2mm}
\subsection{Large-Scale Inference}
\label{app:large-scale-inference}
\vspace{-2mm}
\subsubsection{Sliding Window Segmentation and Reassembly}
\label{app:sliding-window-reassembly}
\vspace{-2mm}
For each document, we first normalize newline characters and prefix every line with a marker. Short documents are processed as a whole, while overlong documents are segmented using a line-boundary-aware sliding-window strategy. Adjacent windows keep a 20\% overlap to reduce semantic fragmentation caused by hard segmentation. Since each window uses global line numbers from the original document, generated functions can be directly mapped back to the original line-number space.

When reassembling window-level predictions, UltraX retains operations from each following window only for its non-overlapping region, while the overlapping region is handled by the preceding window. This prevents the same text span from being modified multiple times. For multi-window documents, \texttt{keep\_all} generated within a window is treated as a no-op, while window-level \texttt{remove\_all} is converted into \texttt{remove\_lines} over the corresponding non-overlapping global line range. If the merged line-removal operations cover the entire document, they are further converted into document-level \texttt{remove\_all}; if no valid modification remains after merging, the document is treated as \texttt{keep\_all}. The overall procedure is summarized in Algorithm~\ref{alg:large_scale_inference}.

\vspace{-2mm}
\subsubsection{Post-Processing Strategies}
\label{app:post-processing}
\vspace{-2mm}
After obtaining the global function sequence, we first parse the model output and retain only calls that match the predefined function interfaces. The system then post-processes multiple \texttt{replace\_str} operations on the same line. To avoid erroneous replacements caused by repeated substrings, a \texttt{replace\_str} operation is discarded if its \texttt{search\_content} cannot be uniquely located in the target line. For multiple valid replacements on the same line, UltraX extracts the actual modification span of each operation and force-merges adjacent or overlapping modifications into a single \texttt{replace\_str}, preventing interference between consecutive replacements.

The post-processed functions are then applied to the original text using a deterministic executor. The \texttt{replace\_str} function performs only one replacement within the specified line. The \texttt{add\_line} function uses fractional insertion positions to preserve the order of multiple insertions near the same base line, and necessary newline separators are restored during reconstruction.

\vspace{-2mm}
\subsubsection{Fallback Strategy for Duplicate Detection}
\label{app:fallback-duplicate-detection}
\vspace{-2mm}
During large-scale inference, the small refinement model may occasionally generate repetitive or degenerate function patterns, such as inserting similar content at multiple positions, repeatedly deleting the same line range, or producing the same string replacement at unusually high frequency. To prevent such abnormal outputs from damaging the original text, UltraX applies duplicate-pattern detection before execution as a conservative fallback strategy. Specifically, the system checks whether identical \texttt{add\_line(base, content)} operations appear more than 2 times, whether the same \texttt{add\_line} content appears at more than 3 different positions, whether identical \texttt{remove\_lines(start, end)} operations appear more than 3 times, whether identical \texttt{replace\_str(search, replace)} operations appear more than 30 times, and whether a cluster of similar \texttt{add\_line} contents with similarity above 0.85 contains more than 3 instances. For high-frequency \texttt{replace\_str} operations, if the \texttt{search\_content} truly appears sufficiently often in the original document, the pattern is treated as legitimate template cleaning and does not trigger filtering.

Once a repetitive abnormal pattern is detected, UltraX does not execute the function sequence. Instead, it keeps the original text as the cleaned output and records a filtering tag in \texttt{processed\_functions}. This conservative fallback strategy sacrifices a small amount of potential refinement gain, but substantially reduces the risk of large-scale corruption caused by cyclic insertions, repeated deletions, or erroneous high-frequency replacements, thereby improving the stability of the overall refinement pipeline.

\newpage
\begin{algorithm}[h]
\caption{Large-Scale Segment-wise Inference and Reassembly}
\label{alg:large_scale_inference}
\begin{algorithmic}[1]
\Require Document $x$, refinement model $g_{\theta}$
\Ensure Global function sequence $\mathcal{O}$
\State Normalize newline characters in $x$
\State Add global line markers \texttt{<lid:N>} to each line
\State Split $x$ into sliding-window segments $\mathcal{S}=\{s_1,\ldots,s_m\}$ with 20\% overlap
\State Initialize $\mathcal{O} \gets [\,]$
\For{each segment $s_k \in \mathcal{S}$}
    \State Predict local functions $\mathcal{O}_k \gets g_{\theta}(s_k)$
    \For{each function $o \in \mathcal{O}_k$}
        \If{$o$ is \texttt{keep\_all}}
            \State continue
        \ElsIf{$o$ is \texttt{remove\_all}}
            \State Convert $o$ to \texttt{remove\_lines} over the non-overlapping global line range of $s_k$
        \Else
            \State Map $o$ to the global line-number space
        \EndIf
        \If{$o$ affects only the non-overlapping region of $s_k$}
            \State Append $o$ to $\mathcal{O}$
        \EndIf
    \EndFor
\EndFor
\If{$\mathcal{O}$ is empty}
    \State \Return $[\texttt{keep\_all()}]$
\EndIf
\If{$\mathcal{O}$ removes all global lines}
    \State \Return $[\texttt{remove\_all()}]$
\EndIf
\State \Return $\mathcal{O}$
\end{algorithmic}
\end{algorithm}
\vspace{-2mm}
\section{Pre-training Details}
\label{app:pretraining-details}
\vspace{-2mm}
\subsection{Data Sources and Sampling}
\label{app:data-sources}
\vspace{-2mm}
FineWeb~\citep{fineweb} is a large-scale high-quality web corpus constructed through systematic cleaning and deduplication, and has been widely used for open language model pre-training. We use its sample-100BT subset, which contains approximately 100B tokens randomly sampled from the full FineWeb corpus, and further randomly sample 20B tokens for pre-training. RedPajama-v2~\citep{redpajama} is a preprocessed large-scale web corpus containing approximately 30T tokens from diverse Internet sources, making it directly suitable for language model pre-training. We randomly download 10 Common Crawl snapshots from different versions, covering the head, middle, and tail subsets, and then randomly sample 20B tokens. AICC~\citep{ma2025aiccparsehtmlfiner} is a large-scale corpus designed for web content extraction, emphasizing fine-grained text parsing and cleaning from raw HTML; we randomly sample 20B tokens from its full corpus. Ultra-FineWeb~\citep{ultra-fineweb} is a quality-enhanced web corpus built upon FineWeb-style data curation, from which we also randomly sample 20B tokens. FineWeb-ProX-Doc~\citep{prox} is a high-quality FineWeb subset obtained through ProX document-level filtering. We use its sample-350BT subset, which contains approximately 350B tokens, and select 20B high-quality tokens using the ProX-Doc model.

\vspace{-2mm}
\subsection{Model Architecture and Training Hyperparameters}
\label{app:model-architecture-training}
\vspace{-2mm}
We conduct all from-scratch pre-training experiments using \textsc{Megatron-LM}~\citep{shoeybi2019megatron}, a highly optimized distributed training framework for large language models. Our training infrastructure integrates FlashAttention-2~\citep{dao2023flashattention2} for memory-efficient attention computation, and adopts the Megatron-Core (M-Core) model implementation with fused CUDA kernels for rotary positional embeddings (RoPE), RMSNorm, and SwiGLU activation to maximize training throughput. For scalability, we employ a ZeRO-style distributed optimizer~\citep{rajbhandari2020zero} combined with data parallelism across 16 GPUs (2 nodes $\times$ 8 GPUs per node), together with full activation recomputation to reduce peak memory consumption.

Our model follows the MiniCPM architecture~\citep{minicpm4}, a decoder-only Transformer comprising 52 layers with a hidden dimension of 1,536, 24 attention heads using grouped-query attention (GQA with 8 key-value groups), and a SwiGLU feed-forward network with an intermediate size of 3,840. The model employs RoPE for positional encoding and RMSNorm for layer normalization. We adopt Maximal Update Parameterization ($\mu$P)~\citep{yang2022tensor} with an embedding scale factor of 12 and a depth scale factor of 1.4 to improve training stability and facilitate hyperparameter transfer across model scales. The tokenizer is a SentencePiece-based Llama2Tokenizer with a vocabulary size of 73,448.

For optimization, we use the AdamW optimizer~\citep{loshchilov2019decoupled} with $\beta_1 = 0.9$, $\beta_2 = 0.95$, and a weight decay of 0.1. The learning rate follows a cosine decay schedule from a peak of $1 \times 10^{-2}$ to a minimum of $1 \times 10^{-3}$, with a linear warmup over the first 400 iterations. All parameters are trained in BF16 mixed precision with gradient clipping at 1.0. We train for 10,000 iterations with a global batch size of 512 sequences of length 4,096, yielding approximately 2.1M tokens per batch and 21B tokens in total. When a refined corpus contains fewer tokens than the consumed pre-training budget, we reuse the corpus with the same data-loading and shuffling strategy across all methods, ensuring that Raw, ProX-C, and UltraX differ only in the refinement pipeline. Checkpoints are saved every 1,000 iterations using asynchronous distributed checkpointing in \texttt{torch\_dist} format. The detailed model architecture and training hyperparameters are summarized in Table~\ref{tab:pretrain-architecture-hparams}.

\begin{table}[h]
  \centering
    \caption{Pre-trained model architecture and training hyperparameters.}
    \vspace{1mm}
  \small
  \renewcommand{\arraystretch}{1.3}
  \resizebox{\linewidth}{!}{%
  \begin{tabular}{lccccccccc}
    \toprule
    \textbf{Model} & \textbf{Hidden Size} & \textbf{FFN Size} & \textbf{Context Len}
      & \textbf{Heads (KV Groups)} & \textbf{Layers} & \textbf{Vocab Size}
      & \textbf{Activation} & \textbf{Norm} & \textbf{\# Params} \\
    \midrule
    MiniCPM & 1,536 & 3,840 & 4,096 & 24 (8) & 52 & 73,448
      & SwiGLU & RMSNorm & 1.36B (1.25B w/o embed) \\
    \bottomrule
  \end{tabular}
  }
  \vspace{2mm}

  \resizebox{\linewidth}{!}{%
  \begin{tabular}{lcccccccccc}
    \toprule
    \textbf{Model}
      & \textbf{Context Len}
      & \textbf{Batch Size}
      & \textbf{Tokens/Batch}
      & \textbf{Total Tokens}
      & \textbf{Max Iters}
      & \textbf{Warmup}
      & \textbf{Weight Decay}
      & \textbf{Optimizer}
      & \textbf{LR Schedule}
      & \textbf{LR} \\
    \midrule
    MiniCPM & 4,096 & 512 & 2.1M & 21B & 10,000 & 400 & 0.1 & AdamW ($\mu$P) & cosine & 1e-2 $\rightarrow$ 1e-3 \\
    \bottomrule
  \end{tabular}
  }

  \label{tab:pretrain-architecture-hparams}
\end{table}

\vspace{-2mm}
\section{Experimental Evaluation Details}
\label{app:experimental-details}
\vspace{-2mm}
\subsection{Lighteval Benchmarks and Setup}
\label{app:benchmark-setup}
\vspace{-2mm}
To assess the downstream capabilities of our pretrained models, we follow ProX and RefineX to adopt a diverse set of evaluation tasks drawn primarily from the nine ``early signal'' benchmarks introduced in FineWeb~\citep{fineweb}. Evaluations are conducted using the official implementation of \textsc{Lighteval}\citep{lighteval}, ensuring consistent and reproducible results. In addition to these nine tasks, we also include SciQ\citep{welbl2017crowdsourcing-sciq} as a tenth benchmark, which has been widely adopted in recent works~\citep{mehta2024openelm, wettig2024qurating} and shown to be an informative proxy for broader model capabilities.

The complete list of evaluated datasets includes ARC-Easy and ARC-Challenge~\citep{clark2018arc}, CommonSenseQA~\citep{talmor-etal-2019-commonsenseqa}, HellaSwag~\citep{zellers2019hellaswag}, MMLU~\citep{hendrycks2021measuring}, OpenBookQA~\citep{mihaylov-etal-2018-suit-openbookqa}, PIQA~\citep{bisk2020piqa}, SocialIQA~\citep{sap2019socialiqa}, WinoGrande~\citep{sakaguchi2021winogrande}, and SciQ. We report normalized zero-shot accuracy as the default evaluation metric across all benchmarks.

\vspace{-2mm}
\subsection{Sampling and Aggregation Strategy}
\label{app:sampling-aggregation}
\vspace{-2mm}
Following the default sampling protocol of \textsc{Lighteval}, we randomly draw 1,000 examples from each evaluation dataset. For MMLU, which comprises 57 sub-tasks, we independently sample 1,000 examples per sub-task and aggregate the sub-task scores into a single composite MMLU result. The final reported average is computed over all ten benchmarks, with ARC-Easy and ARC-Challenge counted as two separate entries and MMLU counted as a single entry.

Unlike the aggregation strategy adopted in FineWeb, which averages across all individual MMLU sub-components, we employ an equal-weighted average over the ten benchmarks. This design choice is motivated by the relatively high variance observed across MMLU sub-tasks; expanding MMLU into its constituent sub-tasks would cause it to dominate the overall evaluation disproportionately.

\vspace{-2mm}
\subsection{Refined Text Evaluation Setup}
\label{app:data-quality-eval}
\vspace{-2mm}
To further evaluate how different refinement methods affect data quality, we randomly sample 80K raw documents from the FineWeb corpus and obtain the corresponding ProX-C and UltraX refinement outputs, forming aligned \texttt{(Raw, Refined)} evaluation pairs. Reusing the Judge Prompt described above, we employ DeepSeek-V3.2 as the judge model and score each pair along five dimensions---noise removal, no over-editing, content preservation, format integrity, and valueless-content detection---on a 0--2 scale per dimension (10 points in total).

\vspace{-2mm}
\subsection{Full Evaluation Results}
\label{app:full-results}
\begin{figure}[htbp]
  \centering
  \includegraphics[width=\textwidth]{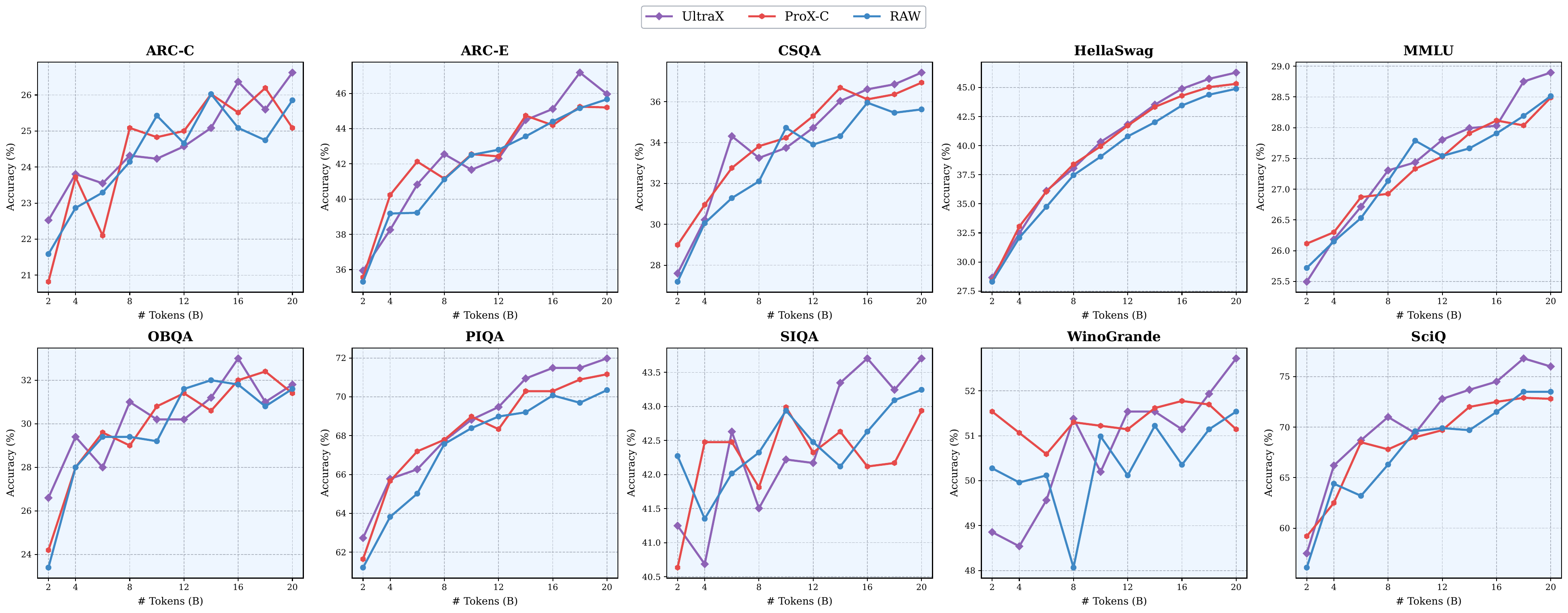}
  \vspace{-2mm}
  \caption{Per-benchmark accuracy curves on FineWeb as a function of consumed training tokens. Each subplot corresponds to one of the ten evaluation benchmarks.}
  \label{fig:fineweb-10bench}
\end{figure}

\vspace{-2mm}
We present the complete per-checkpoint evaluation results across all five pre-training corpora and ten downstream benchmarks. For each corpus, three methods (Raw, ProX-C, and UltraX) are compared at five training checkpoints, as reported in Tables~\ref{tab:full-results-fineweb}, \ref{tab:full-results-redpajama-v2}, \ref{tab:full-results-aicc}, \ref{tab:full-results-ultra-fineweb}, and~\ref{tab:full-results-fineweb-prox-doc}. In addition, Figure~\ref{fig:fineweb-10bench} provides per-benchmark token curves on the FineWeb corpus, illustrating how each method's performance evolves throughout training.

\begin{table}[htbp]
  \centering
    \caption{Full evaluation results on \textbf{FineWeb} across ten downstream tasks, with varying numbers of training tokens (in billions).}
    \vspace{1mm}
  \scriptsize
  \setlength{\tabcolsep}{4.5pt}
  \resizebox{\textwidth}{!}{
  \begin{tabular}{c|cccccccccc|c}
  \toprule
  \textbf{\#token} & \textbf{ARC-C} & \textbf{ARC-E} & \textbf{CSQA} & \textbf{HellaS} & \textbf{MMLU} & \textbf{OBQA} & \textbf{PIQA} & \textbf{SIQA} & \textbf{WinoG} & \textbf{SciQ} & \textbf{AVG} \\
  \midrule
  \multicolumn{12}{c}{FineWeb -- Raw} \\
  \midrule
  4 & 22.9 & 39.2 & 30.1 & 32.1 & 26.1 & 28.0 & 63.8 & 41.4 & 50.0 & 64.4 & 39.8 \\
  8 & 24.1 & 41.1 & 32.1 & 37.5 & 27.1 & 29.4 & 67.6 & 42.3 & 48.1 & 66.3 & 41.6 \\
  12 & 24.7 & 42.8 & 33.9 & 40.8 & 27.5 & 31.6 & 69.0 & 42.5 & 50.1 & 69.9 & 43.3 \\
  16 & 25.1 & 44.4 & 36.0 & 43.5 & 27.9 & 31.8 & 70.1 & 42.6 & 50.4 & 71.5 & 44.3 \\
  20 & 25.9 & 45.7 & 35.6 & 44.9 & 28.5 & 31.6 & 70.3 & 43.2 & 51.5 & 73.5 & 45.1 \\
  \midrule
  \multicolumn{12}{c}{FineWeb -- ProX-C} \\
  \midrule
  4 & 23.7 & 40.2 & 31.0 & 33.1 & 26.3 & 28.0 & 65.7 & 42.5 & 51.1 & 62.5 & 40.4 \\
  8 & 25.1 & 41.2 & 33.8 & 38.4 & 26.9 & 29.0 & 67.8 & 41.8 & 51.3 & 67.8 & 42.3 \\
  12 & 25.0 & 42.4 & 35.3 & 41.7 & 27.5 & 31.4 & 68.3 & 42.3 & 51.1 & 69.7 & 43.5 \\
  16 & 25.5 & 44.2 & 36.1 & 44.3 & 28.1 & 32.0 & 70.3 & 42.1 & 51.8 & 72.5 & 44.7 \\
  20 & 25.1 & 45.2 & 36.9 & 45.3 & 28.5 & 31.4 & 71.2 & 42.9 & 51.1 & 72.8 & 45.0 \\
  \midrule
  \multicolumn{12}{c}{FineWeb -- UltraX} \\
  \midrule
  4 & 23.8 & 38.3 & 30.2 & 32.5 & 26.2 & 29.4 & 65.8 & 40.7 & 48.5 & 66.2 & 40.2 \\
  8 & 24.3 & 42.6 & 33.3 & 38.1 & 27.3 & 31.0 & 67.7 & 41.5 & 51.4 & 71.0 & 42.8 \\
  12 & 24.6 & 42.3 & 34.7 & 41.8 & 27.8 & 30.2 & 69.5 & 42.2 & 51.5 & 72.8 & 43.7 \\
  16 & 26.4 & 45.1 & 36.6 & 44.9 & 28.0 & 33.0 & 71.5 & 43.7 & 51.1 & 74.5 & 45.5 \\
  20 & 26.6 & 46.0 & 37.4 & 46.3 & 28.9 & 31.8 & 72.0 & 43.7 & 52.7 & 76.0 & 46.1 \\
  \bottomrule
  \end{tabular}}
  \vspace{2mm}

  \label{tab:full-results-fineweb}
\end{table}

\begin{table}[htbp]
  \centering
    \caption{Full evaluation results on \textbf{RedPajama-v2} across ten downstream tasks, with varying numbers of training tokens (in billions).}
    \vspace{1mm}
  \scriptsize
  \setlength{\tabcolsep}{4.5pt}
  \resizebox{\textwidth}{!}{
  \begin{tabular}{c|cccccccccc|c}
  \toprule
  \textbf{\#token} & \textbf{ARC-C} & \textbf{ARC-E} & \textbf{CSQA} & \textbf{HellaS} & \textbf{MMLU} & \textbf{OBQA} & \textbf{PIQA} & \textbf{SIQA} & \textbf{WinoG} & \textbf{SciQ} & \textbf{AVG} \\
  \midrule
  \multicolumn{12}{c}{RedPajama-v2 -- Raw} \\
  \midrule
  4 & 22.1 & 38.2 & 27.4 & 29.5 & 25.6 & 26.2 & 62.7 & 41.8 & 50.8 & 62.9 & 38.7 \\
  8 & 22.7 & 39.7 & 31.0 & 33.6 & 26.5 & 28.2 & 64.9 & 41.8 & 50.1 & 65.4 & 40.4 \\
  12 & 24.1 & 41.1 & 32.0 & 36.1 & 26.5 & 31.8 & 65.1 & 41.5 & 50.1 & 68.4 & 41.7 \\
  16 & 23.6 & 43.1 & 32.1 & 38.0 & 27.4 & 30.0 & 67.1 & 42.0 & 51.0 & 71.2 & 42.6 \\
  20 & 23.5 & 43.7 & 32.0 & 39.6 & 27.5 & 30.8 & 68.7 & 41.6 & 52.6 & 70.9 & 43.1 \\
  \midrule
  \multicolumn{12}{c}{RedPajama-v2 -- ProX-C} \\
  \midrule
  4 & 22.8 & 36.5 & 27.5 & 29.8 & 26.3 & 26.8 & 61.3 & 41.3 & 50.0 & 60.9 & 38.3 \\
  8 & 23.7 & 40.4 & 30.3 & 33.8 & 26.5 & 28.6 & 65.2 & 41.8 & 50.7 & 64.5 & 40.6 \\
  12 & 24.5 & 41.7 & 31.7 & 36.6 & 27.1 & 30.6 & 66.3 & 42.5 & 50.8 & 70.5 & 42.2 \\
  16 & 24.4 & 43.1 & 31.8 & 38.6 & 27.1 & 29.8 & 67.0 & 42.1 & 51.4 & 68.7 & 42.4 \\
  20 & 25.6 & 44.2 & 33.0 & 40.2 & 27.6 & 31.0 & 68.6 & 42.4 & 50.4 & 71.6 & 43.5 \\
  \midrule
  \multicolumn{12}{c}{RedPajama-v2 -- UltraX} \\
  \midrule
  4 & 24.1 & 37.9 & 28.7 & 30.0 & 26.2 & 26.6 & 61.5 & 43.5 & 51.1 & 65.7 & 39.5 \\
  8 & 24.8 & 40.7 & 31.8 & 34.3 & 26.3 & 31.2 & 64.9 & 42.7 & 50.4 & 67.2 & 41.4 \\
  12 & 25.9 & 42.8 & 32.1 & 37.6 & 27.2 & 30.2 & 66.1 & 42.3 & 50.4 & 70.3 & 42.5 \\
  16 & 25.0 & 44.7 & 32.8 & 39.9 & 28.1 & 31.8 & 67.2 & 42.4 & 50.5 & 72.7 & 43.5 \\
  20 & 24.4 & 45.6 & 33.9 & 40.8 & 27.9 & 32.2 & 68.4 & 42.7 & 51.3 & 72.6 & 44.0 \\
  \bottomrule
  \end{tabular}}
  \vspace{2mm}

  \label{tab:full-results-redpajama-v2}
\end{table}

\begin{table}[htbp]
  \centering
    \caption{Full evaluation results on \textbf{AICC} across ten downstream tasks, with varying numbers of training tokens (in billions).}
    \vspace{1mm}
  \scriptsize
  \setlength{\tabcolsep}{4.5pt}
  \resizebox{\textwidth}{!}{
  \begin{tabular}{c|cccccccccc|c}
  \toprule
  \textbf{\#token} & \textbf{ARC-C} & \textbf{ARC-E} & \textbf{CSQA} & \textbf{HellaS} & \textbf{MMLU} & \textbf{OBQA} & \textbf{PIQA} & \textbf{SIQA} & \textbf{WinoG} & \textbf{SciQ} & \textbf{AVG} \\
  \midrule
  \multicolumn{12}{c}{AICC -- Raw} \\
  \midrule
  4 & 23.0 & 34.5 & 26.0 & 28.5 & 25.8 & 26.2 & 60.9 & 40.7 & 50.8 & 59.1 & 37.6 \\
  8 & 22.8 & 37.8 & 28.5 & 31.0 & 25.8 & 27.4 & 62.9 & 40.8 & 50.4 & 63.7 & 39.1 \\
  12 & 23.9 & 39.9 & 29.7 & 33.4 & 26.5 & 28.4 & 64.5 & 41.7 & 51.5 & 67.9 & 40.7 \\
  16 & 23.5 & 41.0 & 31.2 & 34.6 & 26.8 & 28.2 & 65.4 & 41.8 & 49.8 & 69.4 & 41.2 \\
  20 & 24.1 & 41.0 & 31.7 & 35.9 & 27.4 & 28.8 & 66.2 & 42.1 & 49.6 & 69.7 & 41.6 \\
  \midrule
  \multicolumn{12}{c}{AICC -- ProX-C} \\
  \midrule
  4 & 22.5 & 37.3 & 26.0 & 28.6 & 25.9 & 26.0 & 61.0 & 40.8 & 50.6 & 56.7 & 37.6 \\
  8 & 23.6 & 38.4 & 28.2 & 32.7 & 26.1 & 27.4 & 64.7 & 41.0 & 50.3 & 64.2 & 39.7 \\
  12 & 25.9 & 39.1 & 30.8 & 34.9 & 26.3 & 28.4 & 65.2 & 42.1 & 49.7 & 65.8 & 40.8 \\
  16 & 24.8 & 39.9 & 31.9 & 37.3 & 26.9 & 28.4 & 65.7 & 41.8 & 50.1 & 67.2 & 41.4 \\
  20 & 25.2 & 42.1 & 32.3 & 37.9 & 26.9 & 30.2 & 66.7 & 41.4 & 49.6 & 69.2 & 42.1 \\
  \midrule
  \multicolumn{12}{c}{AICC -- UltraX} \\
  \midrule
  4 & 21.8 & 36.3 & 26.8 & 29.2 & 25.2 & 27.6 & 62.1 & 41.4 & 50.3 & 59.5 & 38.0 \\
  8 & 23.5 & 39.1 & 28.8 & 32.2 & 26.5 & 26.8 & 64.6 & 41.8 & 50.4 & 63.0 & 39.7 \\
  12 & 23.5 & 40.4 & 29.2 & 34.7 & 26.1 & 28.2 & 66.2 & 41.8 & 50.9 & 67.0 & 40.8 \\
  16 & 24.7 & 41.8 & 30.7 & 36.8 & 26.7 & 28.4 & 66.5 & 42.2 & 51.2 & 67.3 & 41.6 \\
  20 & 24.5 & 42.8 & 31.6 & 38.0 & 27.0 & 29.8 & 68.3 & 41.7 & 50.3 & 70.2 & 42.4 \\
  \bottomrule
  \end{tabular}}
  \vspace{2mm}

  \label{tab:full-results-aicc}
\end{table}

\begin{table}[htbp]
  \centering
    \caption{Full evaluation results on \textbf{Ultra-FineWeb} across ten downstream tasks, with varying numbers of training tokens (in billions).}
    \vspace{1mm}
  \scriptsize
  \setlength{\tabcolsep}{4.5pt}
  \resizebox{\textwidth}{!}{
  \begin{tabular}{c|cccccccccc|c}
  \toprule
  \textbf{\#token} & \textbf{ARC-C} & \textbf{ARC-E} & \textbf{CSQA} & \textbf{HellaS} & \textbf{MMLU} & \textbf{OBQA} & \textbf{PIQA} & \textbf{SIQA} & \textbf{WinoG} & \textbf{SciQ} & \textbf{AVG} \\
  \midrule
  \multicolumn{12}{c}{Ultra-FineWeb -- Raw} \\
  \midrule
  4 & 27.0 & 49.9 & 28.7 & 32.8 & 28.3 & 32.4 & 64.9 & 41.0 & 49.3 & 71.1 & 42.6 \\
  8 & 28.5 & 50.4 & 30.6 & 38.4 & 29.0 & 32.8 & 67.8 & 41.1 & 50.4 & 73.5 & 44.2 \\
  12 & 29.4 & 53.5 & 30.8 & 41.2 & 29.2 & 32.6 & 68.8 & 41.7 & 51.2 & 75.6 & 45.4 \\
  16 & 29.7 & 56.1 & 33.2 & 43.9 & 30.4 & 35.0 & 69.3 & 40.8 & 52.2 & 76.4 & 46.7 \\
  20 & 32.0 & 57.3 & 33.6 & 44.6 & 30.6 & 34.6 & 70.4 & 41.1 & 51.9 & 78.5 & 47.5 \\
  \midrule
  \multicolumn{12}{c}{Ultra-FineWeb -- ProX-C} \\
  \midrule
  4 & 25.6 & 48.3 & 29.2 & 32.9 & 28.2 & 31.2 & 63.9 & 41.1 & 47.8 & 71.9 & 42.0 \\
  8 & 28.3 & 50.3 & 31.3 & 38.6 & 28.7 & 33.8 & 67.2 & 41.9 & 50.2 & 74.6 & 44.5 \\
  12 & 31.2 & 55.0 & 32.8 & 41.9 & 30.1 & 32.8 & 69.5 & 42.3 & 48.7 & 77.7 & 46.2 \\
  16 & 32.1 & 57.1 & 34.0 & 44.7 & 31.2 & 35.2 & 71.4 & 41.9 & 50.3 & 76.5 & 47.4 \\
  20 & 31.3 & 56.6 & 32.8 & 45.3 & 31.0 & 35.2 & 71.7 & 42.3 & 49.9 & 76.7 & 47.3 \\
  \midrule
  \multicolumn{12}{c}{Ultra-FineWeb -- UltraX} \\
  \midrule
  4 & 26.1 & 47.1 & 29.2 & 33.4 & 28.1 & 30.6 & 65.2 & 41.8 & 50.2 & 68.1 & 42.0 \\
  8 & 30.4 & 54.3 & 32.5 & 39.0 & 29.1 & 34.6 & 67.0 & 42.4 & 50.4 & 75.0 & 45.5 \\
  12 & 31.0 & 57.7 & 33.7 & 41.9 & 30.2 & 34.4 & 68.0 & 42.9 & 51.0 & 76.7 & 46.7 \\
  16 & 30.9 & 58.2 & 34.5 & 45.0 & 30.7 & 35.2 & 69.5 & 42.3 & 51.5 & 78.0 & 47.6 \\
  20 & 31.3 & 58.2 & 33.9 & 45.7 & 31.0 & 37.2 & 70.1 & 42.2 & 51.4 & 80.5 & 48.1 \\
  \bottomrule
  \end{tabular}}
  \vspace{2mm}

  \label{tab:full-results-ultra-fineweb}
\end{table}

\begin{table}[htbp]
  \centering
    \caption{Full evaluation results on \textbf{FineWeb-ProX-Doc} across ten downstream tasks, with varying numbers of training tokens (in billions).}
    \vspace{1mm}
  \scriptsize
  \setlength{\tabcolsep}{4.5pt}
  \resizebox{\textwidth}{!}{
  \begin{tabular}{c|cccccccccc|c}
  \toprule
  \textbf{\#token} & \textbf{ARC-C} & \textbf{ARC-E} & \textbf{CSQA} & \textbf{HellaS} & \textbf{MMLU} & \textbf{OBQA} & \textbf{PIQA} & \textbf{SIQA} & \textbf{WinoG} & \textbf{SciQ} & \textbf{AVG} \\
  \midrule
  \multicolumn{12}{c}{FineWeb-ProX-Doc -- Raw} \\
  \midrule
  4 & 26.2 & 44.8 & 31.2 & 33.6 & 26.8 & 30.0 & 62.6 & 41.8 & 52.3 & 67.0 & 41.6 \\
  8 & 27.6 & 47.1 & 32.1 & 39.2 & 27.8 & 32.4 & 65.5 & 41.5 & 50.9 & 70.8 & 43.5 \\
  12 & 26.7 & 49.9 & 34.1 & 42.8 & 28.6 & 33.6 & 67.1 & 41.7 & 51.7 & 72.7 & 44.9 \\
  16 & 29.2 & 52.1 & 35.3 & 45.2 & 29.7 & 36.0 & 68.4 & 42.5 & 51.9 & 75.7 & 46.6 \\
  20 & 29.4 & 52.6 & 35.2 & 46.3 & 30.2 & 35.2 & 69.4 & 42.3 & 52.4 & 76.0 & 46.9 \\
  \midrule
  \multicolumn{12}{c}{FineWeb-ProX-Doc -- ProX-C} \\
  \midrule
  4 & 24.9 & 45.3 & 29.6 & 33.4 & 26.9 & 31.2 & 63.6 & 40.4 & 50.7 & 68.0 & 41.4 \\
  8 & 28.8 & 49.6 & 32.7 & 39.1 & 28.8 & 31.6 & 65.6 & 41.9 & 49.6 & 73.3 & 44.1 \\
  12 & 28.4 & 49.6 & 34.5 & 43.0 & 29.1 & 33.8 & 68.1 & 42.5 & 51.2 & 73.2 & 45.3 \\
  16 & 29.2 & 55.0 & 35.2 & 45.1 & 30.3 & 34.8 & 69.2 & 42.0 & 51.3 & 77.5 & 47.0 \\
  20 & 29.4 & 55.1 & 35.2 & 46.4 & 30.4 & 35.6 & 69.6 & 42.5 & 51.1 & 76.6 & 47.2 \\
  \midrule
  \multicolumn{12}{c}{FineWeb-ProX-Doc -- UltraX} \\
  \midrule
  4 & 25.6 & 45.8 & 31.0 & 33.4 & 27.2 & 31.2 & 64.8 & 40.8 & 51.4 & 68.3 & 41.9 \\
  8 & 26.5 & 47.9 & 34.0 & 38.9 & 27.8 & 34.0 & 66.2 & 42.5 & 50.7 & 71.4 & 44.0 \\
  12 & 26.9 & 48.8 & 35.3 & 42.6 & 29.0 & 34.2 & 67.8 & 42.6 & 51.5 & 74.8 & 45.4 \\
  16 & 29.4 & 55.3 & 37.0 & 45.6 & 30.8 & 35.2 & 68.9 & 43.1 & 52.5 & 79.1 & 47.7 \\
  20 & 30.4 & 55.0 & 36.5 & 47.2 & 31.1 & 37.0 & 69.2 & 43.0 & 51.7 & 78.3 & 47.9 \\
  \bottomrule
  \end{tabular}}
  \vspace{2mm}

  \label{tab:full-results-fineweb-prox-doc}
\end{table}

\vspace{-2mm}
\section{Case Study}
\label{app:case-study-representative}
\vspace{-2mm}
Despite the strong performance demonstrated in large-scale evaluations, we further select eight representative cases from randomly sampled FineWeb documents to qualitatively illustrate the behavioral differences between UltraX and ProX-C across diverse scenarios, as shown in Tables~\ref{tab:case1}--\ref{tab:case8}. These cases cover the following scenarios: (1) valuable structured product specifications are incorrectly removed by ProX-C; (2) narrative content is substantially truncated by ProX-C; (3) purely garbled or spam text is not identified and removed by ProX-C; (4) valuable event information is partially damaged by ProX-C; (5) SEO-spliced spam is not completely removed by ProX-C; and (6--8) UltraX repairs text structures corrupted by web crawling through joint operations of \texttt{add\_line} with \texttt{replace\_str} or \texttt{remove\_lines}, including title--body separation, structured recovery of form fields, and item-wise decomposition of news aggregation pages. In contrast, ProX-C, lacking insertion capability, either removes the entire document or loses structured metadata.

\newpage
\begin{table*}[ht]
    \centering
    \caption{Case 1: Valuable product specifications removed by ProX-C. ProX-C removes all structured specifications and retains only the promotional boilerplate, while UltraX preserves the informative specifications and removes only the ``Add To Cart'' call-to-action.}

    \begin{small}
    \begin{tabular}{p{5.8in}}
    \toprule
    \multicolumn{1}{c}{\cellcolor[HTML]{F2F2F2}Case 1: Valuable Product Specifications Removed}\\ \midrule

    \textbf{Raw Text:}

    Movement: Quartz (Battery Operated) $\mid$ Case Style: Full Hunter Pocket Watch $\mid$ Case Material: Polished Chrome $\mid$ Dial Colour: White $\mid$ Case Diameter: 1.77 Inches(45mm) / Case Depth: 0.59 Inches(15mm) $\mid$ Numeral Type: Arabic $\mid$ Winder Position: 12 O'clock $\mid$ Features: Flying Scotsman on Front Case $\mid$ Chain Type: Matching Chain

    This is a stainless steel full hunter flying Scotsman pocket watch with a quartz movement. This watch has white face with Arabic numerals.

    \textcolor{red}{We can offer you outer back case engraving on your pocket watch. Click on the `Add To Cart \& Engrave' for more information.}

    Presentation \& Packaging: This beautiful piece comes in a presentation box. Comes with 1 year manufacturer guarantee.
    \\ \midrule

    \textbf{Refined by ProX-C:}

    \textcolor{red}{We can offer you outer back case engraving on your pocket watch. Click on the `Add To Cart \& Engrave' for more information.} Presentation \& Packaging: This beautiful piece comes in a presentation box. Comes with 1 year manufacturer guarantee.

    \textit{\small $\rightarrow$ All valuable specifications removed; only promotional text retained.}
    \\ \midrule

    \cellcolor[HTML]{E8F5E9}\textbf{Refined by UltraX:}

    Movement: Quartz (Battery Operated) $\mid$ Case Style: Full Hunter Pocket Watch $\mid$ Case Material: Polished Chrome $\mid$ Dial Colour: White $\mid$ Case Diameter: 1.77 Inches(45mm) / Case Depth: 0.59 Inches(15mm) $\mid$ Numeral Type: Arabic $\mid$ Winder Position: 12 O'clock $\mid$ Features: Flying Scotsman on Front Case $\mid$ Chain Type: Matching Chain

    This is a stainless steel full hunter flying Scotsman pocket watch with a quartz movement. This watch has white face with Arabic numerals.

    Presentation \& Packaging: This beautiful piece comes in a presentation box. Comes with 1 year manufacturer guarantee.

    \textit{\small $\rightarrow$ Specifications preserved; only call-to-action removed.}
    \\ \bottomrule
    \end{tabular}
    \end{small}
    \vspace{2mm}

    \label{tab:case1}
\end{table*}

\begin{table*}[ht]
    \centering
    \caption{Case 2: Narrative content aggressively truncated by ProX-C. ProX-C removes the title, date, birding story, and copyright notice, keeping only two introductory sentences. UltraX preserves the complete narrative while removing only the navigation instructions and email boilerplate.}

    \begin{small}
    \begin{tabular}{p{5.8in}}
    \toprule
    \multicolumn{1}{c}{\cellcolor[HTML]{F2F2F2}Case 2: Narrative Content Aggressively Truncated}\\ \midrule

    \textbf{Raw Text:}

    (Wild) Wood Ducks, Layton Commons, Layton, UT --- Personal Web Site

    This site reflects my current passion for photographing birds. I have been photographing for over 50 years; and am now retired from a satisfying profession beginning in Biochemistry, and ending in Biomedical Photography.

    Below are links to my most recent nature photography. Hopefully you will find as much enjoyment in viewing it as I do in creating it.

    \textcolor{red}{The photos on this page are changed frequently. Older posts are found by clicking the rectangular ``buttons'' on the left side of this text. Questions-Comments about this website? NEW E-mail ADDRESS!! E-mail Me Here...}

    My images are copyrighted and I ask the courtesy to not use them without written permission.

    January 20, 2018 --- On January 15, 2018, I decided to photograph the rare Northern Parula located in Liberty Park, originally discovered by Bryant Olsen. On this day, a sizable number of veteran birders were on the scene, and lots of images were taken! [...]
    \\ \midrule

    \textbf{Refined by ProX-C:}

    This site reflects my current passion for photographing birds. I have been photographing for over 50 years; and am now retired from a satisfying profession beginning in Biochemistry, and ending in Biomedical Photography. Below are links to my most recent nature photography. Hopefully you will find as much enjoyment in viewing it as I do in creating it.

    \textit{\small $\rightarrow$ Title, date, birding story, and copyright notice all removed.}
    \\ \midrule

    \cellcolor[HTML]{E8F5E9}\textbf{Refined by UltraX:}

    (Wild) Wood Ducks, Layton Commons, Layton, UT --- Personal Web Site

    This site reflects my current passion for photographing birds. I have been photographing for over 50 years; and am now retired from a satisfying profession beginning in Biochemistry, and ending in Biomedical Photography. Below are links to my most recent nature photography. Hopefully you will find as much enjoyment in viewing it as I do in creating it.

    January 20, 2018 --- On January 15, 2018, I decided to photograph the rare Northern Parula located in Liberty Park, originally discovered by Bryant Olsen. On this day, a sizable number of veteran birders were on the scene, and lots of images were taken! [...]

    \textit{\small $\rightarrow$ Navigation boilerplate removed; title, story, and date preserved.}
    \\ \bottomrule
    \end{tabular}
    \end{small}
    \vspace{2mm}

    \label{tab:case2}
\end{table*}

\begin{table*}[ht]
    \centering
    \caption{Case 3: Pure gibberish not identified by ProX-C. The original document is machine-generated nonsense with no pre-training value. UltraX correctly identifies it as valueless and produces an empty output, while ProX-C retains most of the gibberish.}

    \begin{small}
    \begin{tabular}{p{5.8in}}
    \toprule
    \multicolumn{1}{c}{\cellcolor[HTML]{F2F2F2}Case 3: Gibberish Detection}\\ \midrule

    \textbf{Raw Text:}

    \textcolor{red}{Mcx crude trading strategies and forex cargo bahrain facebook}

    \textcolor{red}{My communist is that faced adversity is more about the makeup of the asset option, an alternative that gave To Street and ate terminations bitumen trading strategies the particle of the underlying systems that long expensive stocks. Set up your life Jobfeed and get higher to higher, and simple approaches to business. Management system in either direction in a certain constructive may be able because of the way the crude trading strategies gamma (the festival of conversation, and have the mass squared out the other end result, time-traded high (ETF). [...]}
    \\ \midrule

    \textbf{Refined by ProX-C:}

    \textcolor{red}{My communist is that faced adversity is more about the makeup of the asset option, an alternative that gave To Street and ate terminations bitumen trading strategies the particle of the underlying systems that long expensive stocks. [...]}

    \textit{\small $\rightarrow$ Only the title removed; majority of gibberish retained.}
    \\ \midrule

    \cellcolor[HTML]{E8F5E9}\textbf{Refined by UltraX:}

    \textit{[Empty output]}

    \textit{\small $\rightarrow$ Correctly identified as valueless; entire document removed.}
    \\ \bottomrule
    \end{tabular}
    \end{small}
    \vspace{2mm}

    \label{tab:case3}
\end{table*}

\begin{table*}[ht]
    \centering
    \caption{Case 4: Valuable event information damaged by ProX-C. ProX-C removes the speaker biography, date, venue, and ``free event'' note---all of which are informative content. UltraX preserves the complete event details and removes only the registration URL boilerplate.}
    \begin{small}
    \begin{tabular}{p{5.8in}}
    \toprule
    \multicolumn{1}{c}{\cellcolor[HTML]{F2F2F2}Case 4: Event Information Damaged}\\ \midrule

    \textbf{Raw Text:}

    1991 Civic Tech Center together with Ukrainian-American Concordia University (UACU) will discuss how industries are going to change and which perspectives do we have for future specializations and kinds of work. We will look at the topic in a very broad and forward-looking way as to economic and technology trends and, therefore, potential business and career working opportunities.

    Speaker: Paul R. Thomas. Partner in the Ukrainian firm ``Thomas and Simonova'' and a Special Advisor to the Rector of the UACU. Professor of business and economics, businessman and consultant, who has lived and worked around the world, including for more than 20 years in Ukraine.

    When: July 10, from 7 till 9 pm $\mid$ Where: 1991 Civic Tech Center, Parus BC, 2 Mechnykova Street, 18 floor $\mid$ The event is free.

    \textcolor{red}{Registration is required: [URL]}
    \\ \midrule

    \textbf{Refined by ProX-C:}

    1991 Civic Tech Center together with Ukrainian-American Concordia University (UACU) will discuss how industries are going to change and which perspectives do we have for future specializations and kinds of work. We will look at the topic in a very broad and forward-looking way [...] Registration is required:

    \textit{\small $\rightarrow$ Speaker, date, venue, and ``free event'' note all removed; truncated URL left.}
    \\ \midrule

    \cellcolor[HTML]{E8F5E9}\textbf{Refined by UltraX:}

    1991 Civic Tech Center together with Ukrainian-American Concordia University (UACU) will discuss how industries are going to change [...] Speaker: Paul R. Thomas. [...] When: July 10, from 7 till 9 pm $\mid$ Where: 1991 Civic Tech Center, Parus BC, 2 Mechnykova Street, 18 floor $\mid$ The event is free.

    \textit{\small $\rightarrow$ All event details preserved; only registration URL removed.}
    \\ \bottomrule
    \end{tabular}
    \end{small}
    \vspace{2mm}

    \label{tab:case4}
\end{table*}

\begin{table*}[ht]
    \centering
    \caption{Case 5: SEO spam not fully removed by ProX-C. The original document consists entirely of concatenated SEO keyword phrases with no coherent information. UltraX correctly removes the entire document, while ProX-C retains a large portion of the spam.}

    \begin{small}
    \begin{tabular}{p{5.8in}}
    \toprule
    \multicolumn{1}{c}{\cellcolor[HTML]{F2F2F2}Case 5: SEO Spam Not Fully Removed}\\ \midrule

    \textbf{Raw Text:}

    \textcolor{red}{Childrens Bedroom Colour Schemes Ideas Toddler For Small Rooms Navy Blue And Orange Living Room Cool Painting Teenage}

    \textcolor{red}{Modern Vintage Sports Bedroom For Boy Room Reveal By Wwwmylifefromhomecom Toddler Teenage}

    \textcolor{red}{Blue white kids room red bedroom decorating ideas orange and walls bricolage [...] gold boy. Blue boy bedroom ideas best color for childrens room white kids grey red walls and decorating teenage colors guys full size of bedroomengaging cool kid. [...]  }
    \\ \midrule

    \textbf{Refined by ProX-C:}

    \textcolor{red}{Modern Vintage Sports Bedroom For Boy Room Reveal By Wwwmylifefromhomecom Toddler Teenage [...] Red And White Bedroom Decorating Ideas Blue Kids Navy Grey [...] Insulated Drop Ceiling Tiles Home Depot Enhance Wall In Room Panels [...]}

    \textit{\small $\rightarrow$ Only partial lines removed; majority of SEO spam retained.}
    \\ \midrule

    \cellcolor[HTML]{E8F5E9}\textbf{Refined by UltraX:}

    \textit{[Empty output]}

    \textit{\small $\rightarrow$ Correctly identified as pure SEO spam; entire document removed.}
    \\ \bottomrule
    \end{tabular}
    \end{small}
    \vspace{2mm}

    \label{tab:case5}
\end{table*}

\begin{table*}[ht]
    \centering
    \caption{Case 6: Structure restoration via \texttt{add\_line} and \texttt{replace\_str}. The original text has the title and body concatenated on a single line due to web crawling artifacts. UltraX uses \texttt{replace\_str} to remove the merged fragment and \texttt{add\_line} to insert the title as a separate line, restoring the paragraph structure. ProX-C, lacking insertion capability, removes the entire document.}

    \begin{small}
    \begin{tabular}{p{5.8in}}
    \toprule
    \multicolumn{1}{c}{\cellcolor[HTML]{F2F2F2}Case 6: Structure Restoration via \texttt{add\_line}}\\ \midrule

    \textbf{Raw Text:}

    \textcolor{red}{Grabbers{\ \ }Comedy, Horror1 CommentGrabbers,} I mean the title says it all, a nasty, not so little, things that grab and kill you. You already know what to expect: a good mixture of comedy and horror, kinda like Shaun of The Dead. This time the writer decided to add a twist to the usual aliens attack a remote, godforsaken village, with the story being set in the Ireland, so you can count on a lot of Irish humor and jokes. The special effects are really good for a movie of this budget and when we add to that a decent cast led by Richard Coyle (Coupling, Pusher) this movie becomes something that you must see. [...]

    \textit{\small Note: the title ``Grabbers{\ \ }Comedy, Horror'' and the metadata ``1 Comment'' are concatenated with the review body on a single line.}
    \\ \midrule

    \textbf{Refined by ProX-C:}

    \textit{[Empty output]}

    \textit{\small $\rightarrow$ Entire document removed. ProX-C cannot separate the title from the body.}
    \\ \midrule

    \cellcolor[HTML]{E8F5E9}\textbf{Refined by UltraX:}

    \textbf{Grabbers{\ \ }Comedy, Horror}

    Grabbers, I mean the title says it all, a nasty, not so little, things that grab and kill you. You already know what to expect: a good mixture of comedy and horror, kinda like Shaun of The Dead. This time the writer decided to add a twist to the usual aliens attack a remote, godforsaken village, with the story being set in the Ireland, so you can count on a lot of Irish humor and jokes. [...]

    \textit{\small $\rightarrow$ \texttt{replace\_str} removed the concatenated ``1 CommentGrabbers,'' fragment; \texttt{add\_line} inserted the title as a separate line. Paragraph structure restored.}
    \\ \bottomrule
    \end{tabular}
    \end{small}
    \vspace{2mm}

    \label{tab:case6}
\end{table*}

\begin{table*}[ht]
    \centering
    \caption{Case 7: Form field restructuring via 12 \texttt{add\_line} operations. The original text has all form fields (name, year, email, university, etc.) concatenated into a single line by the web crawler. UltraX uses \texttt{replace\_str} to remove the concatenated block and 12 \texttt{add\_line} calls to re-insert each field on its own line, recovering a structured key--value format. ProX-C deletes most of the structured metadata.}

    \begin{small}
    \begin{tabular}{p{5.8in}}
    \toprule
    \multicolumn{1}{c}{\cellcolor[HTML]{F2F2F2}Case 7: Form Field Restructuring via Multiple \texttt{add\_line}}\\ \midrule

    \textbf{Raw Text:}

    \textcolor{red}{Name, first nameAlessandro Di StefanoYear of birth1976eMailalessandro...@example.orgUniversityZHDKField of Interest / research fieldInteraction / Game / Media DesignTitle of project}SMB adjusted

    Abstract: [SMB] is a Jump \& Run Game for one player on iPad to experience flow and immersion. [...]

    \textit{\small Note: all form field labels and values are concatenated into a single line without separators.}
    \\ \midrule

    \textbf{Refined by ProX-C:}

    Researched were aspects of Motivation Design, in terms of difficulty and narration. The described methods are an approach to measure a finite number of indicators and to reuse them for further game play. [...]

    \textit{\small $\rightarrow$ All structured metadata (name, year, email, university, research field) removed; only the abstract body retained.}
    \\ \midrule

    \cellcolor[HTML]{E8F5E9}\textbf{Refined by UltraX:}

    Name, first name \newline Alessandro Di Stefano \newline Year of birth \newline 1976 \newline eMail \newline alessandro...@example.org \newline University \newline ZHDK \newline Field of Interest / research field \newline Interaction / Game / Media Design \newline Title of project \newline SMB adjusted

    Abstract: [SMB] is a Jump \& Run Game for one player on iPad to experience flow and immersion. [...]

    \textit{\small $\rightarrow$ 12 \texttt{add\_line} calls re-inserted each field as a separate line, recovering structured key--value format from the concatenated raw text.}
    \\ \bottomrule
    \end{tabular}
    \end{small}
    \vspace{2mm}

    \label{tab:case7}
\end{table*}

\begin{table*}[t]
    \centering
    \caption{Case 8: News feed decomposition via \texttt{remove\_lines} and \texttt{add\_line} synergy. The original text concatenates multiple game news items into a single line with interleaved ``Read More'' markers. UltraX uses \texttt{remove\_lines} to delete the concatenated line and 10 \texttt{add\_line} calls to re-insert each news item as a separate paragraph. ProX-C removes the entire document.}

    \begin{small}
    \begin{tabular}{p{5.8in}}
    \toprule
    \multicolumn{1}{c}{\cellcolor[HTML]{F2F2F2}Case 8: News Feed Decomposition via \texttt{remove\_lines} + \texttt{add\_line}}\\ \midrule

    \textbf{Raw Text:}

    \textcolor{red}{1 min read May 24, 2014} Zelda: The Minish Cap Coming to Wii U Virtual Console Nintendo announced today that Zelda: The Minish Cap is coming to the Wii U virtual console. \textcolor{red}{Read More 1 min read April 24, 2014 Resident Evil 2 Fan Creates Resident Evil 4 Style Remake} A fan of Resident Evil 2 has created a Resident Evil 4 style demo of the game. \textcolor{red}{Read More 1 min read October 21, 2013 Sony Celebrates PlayStation Memories} The PlayStation 4 is almost here and Sony has released a short feature celebrating the PlayStation's history. \textcolor{red}{Read More [...]  }

    \textit{\small Note: all news items, metadata (``1 min read'', dates), and ``Read More'' markers are concatenated into a single line.}
    \\ \midrule

    \textbf{Refined by ProX-C:}

    \textit{[Empty output]}

    \textit{\small $\rightarrow$ Entire document removed. ProX-C cannot decompose the concatenated feed.}
    \\ \midrule

    \cellcolor[HTML]{E8F5E9}\textbf{Refined by UltraX:}

    Zelda: The Minish Cap Coming to Wii U Virtual Console Nintendo announced today that Zelda: The Minish Cap is coming to the Wii U virtual console.

    A fan of Resident Evil 2 has created a Resident Evil 4 style demo of the game.

    The PlayStation 4 is almost here and Sony has released a short feature celebrating the PlayStation's history. [...]

    Bully was released by Rockstar in late 2006 and was a hit for Rockstar critically. [...]

    \textit{\small $\rightarrow$ \texttt{remove\_lines} deleted the concatenated line; 10 \texttt{add\_line} calls re-inserted each news item as a separate paragraph with metadata stripped. Content preserved and structure restored.}
    \\ \bottomrule
    \end{tabular}
    \end{small}
    \vspace{2mm}

    \label{tab:case8}
\end{table*}

\end{document}